\documentclass{article}

\usepackage[utf8]{inputenc} 
\usepackage[T1]{fontenc}    
\usepackage{hyperref}       
\usepackage{url}            
\usepackage{booktabs}       
\usepackage{amsfonts}       
\usepackage{nicefrac}       
\usepackage{microtype}      
\usepackage{xcolor}         
\usepackage{graphicx}
\usepackage{subcaption}
\usepackage{array}
\usepackage{cancel}
\usepackage{pifont}    
\usepackage{multirow}
\usepackage{hyperref}

\usepackage{tikz}
\usetikzlibrary{shadings}

\pgfdeclarehorizontalshading{jetColormap}{0.5cm}{
 color(0cm)=(blue); 
 color(0.4cm)=(cyan); 
 color(0.5cm)=(green);
 color(0.6cm)=(yellow); 
 color(0.7cm)=(orange); 
 color(1cm)=(red) 
}

\newcommand{\kcu}{\mathbf{k}_{\cdot\mathbf{u}}}
\newcommand{\kuc}{\mathbf{k}_{\mathbf{u}\cdot}}
\newcommand{\kuu}{\mathbf{K}_{\bf uu}}
\newcommand{\kcc}{k(\cdot, \cdot)}

\newcommand{\colorbar}{
\begin{tikzpicture}[baseline={(0,.15)}]
    \shade[shading=jetColormap] (0,0) rectangle (1,.5);
\end{tikzpicture}}

\newcommand{\customcaption}[2]{
    \caption[#1]{#1\colorbar #2}
}

\newcommand{\xmark}{\ding{55}}

\usepackage[accepted]{icml2025}

\usepackage{amsmath}
\usepackage{amssymb}
\usepackage{mathtools}
\usepackage{amsthm}

\usepackage[capitalize,noabbrev]{cleveref}

\theoremstyle{plain}

\newtheorem*{guarantee}{Guarantee}

\theoremstyle{definition}

\theoremstyle{remark}

\usepackage[textsize=tiny]{todonotes}

\icmltitlerunning{Adjusting Model Size in Continual Gaussian Processes: How Big is Big Enough? }

\begin{document}

\twocolumn[
\icmltitle{Adjusting Model Size in Continual Gaussian Processes: \\ How Big is Big Enough?}



\icmlsetsymbol{equal}{*}

\begin{icmlauthorlist}
\icmlauthor{Guiomar Pescador-Barrios}{imperial}
\icmlauthor{Sarah Filippi}{imperial}
\icmlauthor{Mark van der Wilk}{oxford}
\end{icmlauthorlist}

\icmlaffiliation{imperial}{Department of Mathematics, Imperial College London, UK}
\icmlaffiliation{oxford}{Department of Computer Science, University of Oxford, UK}
\icmlcorrespondingauthor{Guiomar Pescador-Barrios}{glp22@ic.ac.uk}

\icmlkeywords{Machine Learning, ICML, continual learning, gaussian processes, adaptive model capacity, variational inference, inducing points, sparse approximations, bayesian nonparametrics}

\vskip 0.3in
]



\printAffiliationsAndNotice{}  

\begin{abstract}
Many machine learning models require setting a parameter that controls their size before training, e.g.~number of neurons in DNNs, or inducing points in GPs. Increasing capacity typically improves performance until all the information from the dataset is captured. After this point, computational cost keeps increasing, without improved performance. This leads to the question ``How big is big enough?'' 
We investigate this problem for Gaussian processes (single-layer neural networks) in continual learning. Here, data becomes available incrementally, and the final dataset size will therefore not be known before training, preventing the use of heuristics for setting a fixed model size. We develop a method to automatically adjust model size while maintaining near-optimal performance. 
Our experimental procedure follows the constraint that any hyperparameters must be set without seeing dataset properties, and we show that our method performs well across diverse datasets without the need to adjust its hyperparameter, showing it requires less tuning than others.

\end{abstract}
\section{Introduction}

\begin{figure*}[ht]
\vskip 0.2in
\begin{center}
\centerline{\includegraphics[width=0.9\linewidth]{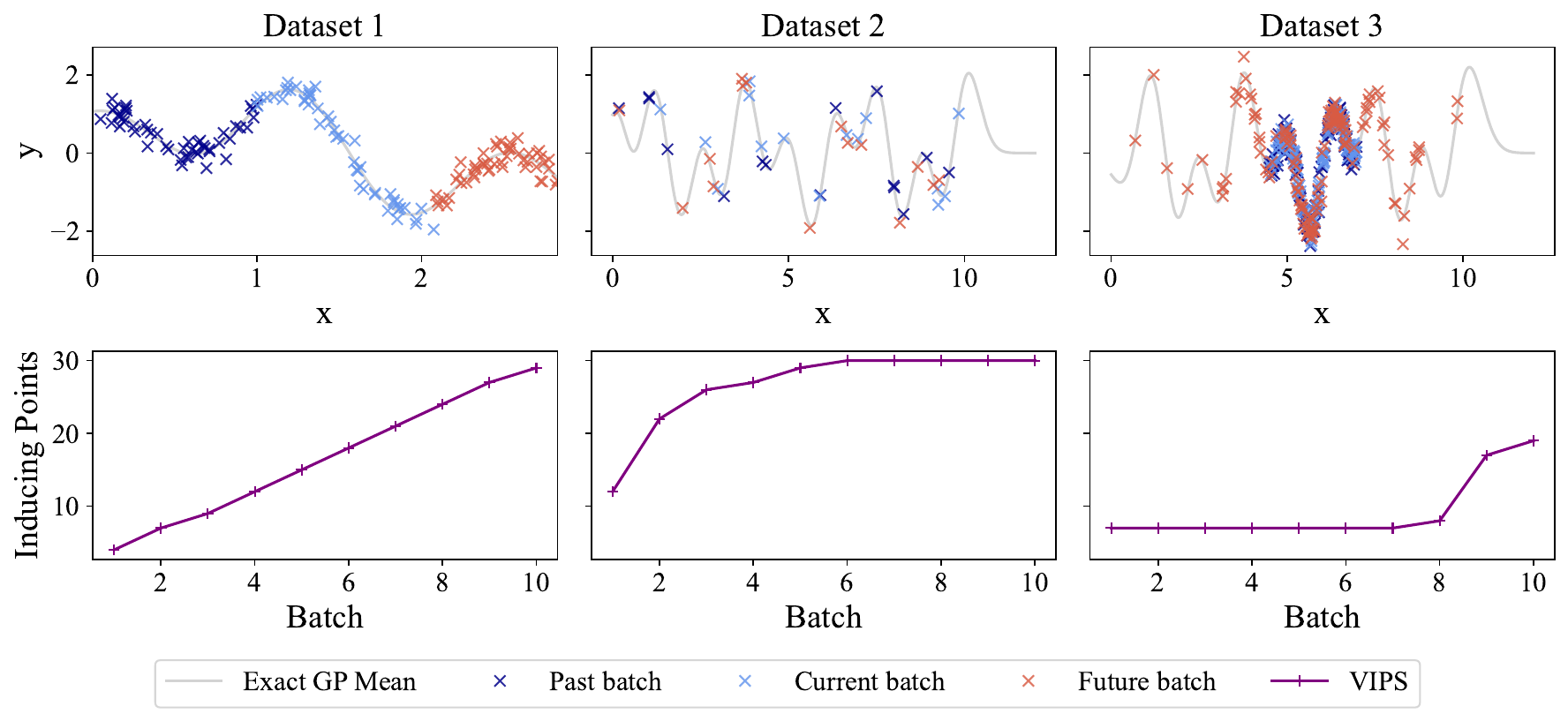}}
\caption{Three continual learning scenarios with different capacity requirements. \textbf{Top}: Three consecutive batches for 1) a growing input space, 2) i.i.d.~samples from a uniform distribution, and 3) narrow-range samples with occasional outliers. 
\textbf{Bottom}: Number of inducing points selected using the VIPS algorithm at each batch. When selecting model size with our method, we observed: 1) a linear increase, 2) after initial training, we see a halt in growth, and 3) a low model size until it encounters outliers.}
\label{fig:data_types}
\end{center}
\vskip -0.2in
\end{figure*}

Continual learning aims to train models when the data arrives in a stream of batches, without storing data after it has been processed, and while obtaining predictive performance that is as high as possible at each point in time \cite{ring1997child}. Selecting the size of the model is challenging in this setting, since typical non-continual training procedures do this by trial-and-error (cross-validation) using repeated training runs, which is not possible under our requirement of not storing any data.
This is a crucial parameter to set well, since if the model is too small, predictive performance will suffer. One solution could be to simply make all continual learning models so large, that they will always have enough capacity, regardless of what dataset and what amount of data they will be given. However, this ``worst-case'' strategy is wasteful of computational resources.

A more elegant solution would be to grow the size of the model adaptively as data arrives, according to the needs of the problem (see Figure \ref{fig:data_types} for an illustration). For example, if data were only ever gathered from the same region, there would be diminishing novelty in every new batch. In this case, one would want the model size to stabilise, with growth resuming once data arrives from new regions. In this paper, we investigate a principle for determining how to select the size of a model so that it is sufficient to obtain near-optimal performance, while otherwise wasting a minimal amount of computation. In other words, we seek to answer the question of ``how big is big enough?'' for setting the size of models throughout continual learning.

We investigate this question for Gaussian processes, since excellent continual learning methods exist that perform very similarly to full-batch methods, but which assume a fixed model capacity that is large enough. We provide a criterion for determining the number of inducing variables (analogous to neurons) that are needed whenever a new batch of data arrives. 
We show that our method is better able to maintain performance close to optimal full-batch methods, with a slower growth of computational resources compared to other continual methods, which either make models too small and perform poorly, or too large and waste computation. 
One hyperparameter needs to be tuned to control the balance between computational cost and accuracy. However, a single value works similarly across datasets with different properties, allowing all modelling decisions to be made before seeing any data.

Our approach benefits from Bayesian nonparametric perspectives, by separating the specification of the capacity of the model, which should be able to learn from an unbounded amount of data, from the specification of the approximation, which determines how much computational effort is needed to represent the solution for the current finite dataset. Our approach relies on the variational inducing variable approximation of \citet{Titsias2009b}, which leads to an interpretation that our method automatically adjusts the width of a single-layer neural network. Our hope is that such perspectives will be useful for the development of adaptive deep neural networks, which would allow models to be trained on large amounts of data beyond the limitations of storage infrastructure, or learning from streams of data that come directly from interaction with an environment.

\section{Related Work}

The foundational problem that makes training with gradient descent insufficient for continual learning is \emph{catastrophic forgetting}, where previously acquired knowledge is lost in favour of recent information \citep{mccloskey1989catastrophic,goodfellow2013empirical}. Many solutions have been proposed in the literature, such as encouraging weights to be close to values that were well-determined by past data \citep{kirkpatrick2017overcoming, schwarz2018progress}, storing subsets or statistics of past data to continue to train the neural network in the future \citep{li2017learning,lopez2017gradient}, and approximate Bayesian methods that balance uncertainty estimates of parameters with the strength of the data \citep{Bui2017,Nguyen2018,rudner22functional,chang2023low}. Within continual learning, many different settings have been investigated, which vary in difficulty \citep{farquhar2018towards}. Across these tasks, the gap in performance to a full-batch training procedure therefore also varies, but despite progress, some gap in performance remains.

Bayesian continual learning methods have been developed because the posterior given past data becomes the prior for future data, making the posterior a sufficient quantity to estimate \citep[\S 19.7]{pml2onlineblr}. For the special case of linear-in-the-parameters regression models, the posterior and updates can be calculated in closed form, leading to continual learning giving \emph{exactly} the same result as full-batch training. In most cases (e.g.~for neural networks), the posterior cannot be found exactly, necessitating the aforementioned approximations that use an approximate posterior as the sufficient quantity \citep{Nguyen2018,rudner22functional,chang2023low}.

Even with a perfect solution to catastrophic forgetting (e.g.~in the case of linear-in-the-parameters regression models), continual learning methods face the additional difficulty of ensuring that models have sufficient capacity to accommodate the continuously arriving information. In continual learning, it is particularly difficult to determine a fixed size for the model, since the number of data or tasks are not yet known, and selecting a model that is too small can significantly hurt performance. Growing models with data can fix this problem. For example, \citet{rusu2016progressive} extend hidden representations by a fixed amount for each new batch. \citet{jae2018lifelong} argue that extension by a fixed amount is wasteful and should instead be data dependent, specifically by copying neurons if their value changes too much, and adding new neurons if the training loss doesn't reach a particular threshold.  \citet{kessler2021hierarchical} propose to use the Indian Buffet Process as a more principled way to regularise how fast new weights are added with tasks. \citet{maile2022and,mitchell2023self} develop similar ideas of automatically adjusting model size to the normal batch-learning setting.
We agree with the aims of these papers, but note that they rely on thresholds that need to be tuned to the dataset characteristics, e.g.~tolerances on the training error which depend on the amount of intrinsic noise in the dataset. We seek a unifying training principle that is broadly applicable to a wide range of datasets, without knowing their properties a priori.

Growing model capacity with dataset size was one of the main justifications for research into (Bayesian) non-parametric models \citep{ghahramani2013nonparametrics,ghahramani2015probabilistic}. This approach defines models with infinite capacity, with Bayesian inference naturally using an appropriate finite capacity to make predictions, with finite compute budgets. Gaussian processes (GPs) \citep{RasmussenWilliams2005} are the most common Bayesian non-parametric model for supervised learning, and are equivalent to infinitely-wide deep neural networks \citep{neal1996bayesian,matthews2018wide} and linear-in-the-parameters models with an infinite feature space \citep{mercer1909}. Their infinite capacity allows them to recover functions perfectly in the limit of infinite data \citep{vdvaart2008rates}, and their posterior can be computed in closed form. These are strong principles for providing high-quality solutions to both catastrophic forgetting and ensuring appropriate capacity, and therefore make GPs an excellent model for studying continual learning. 

However, developing a practical continual learning in GPs is not as straightforward as it is in finite dimensional linear models, because (for $N$ datapoints) the posterior requires \textbf{1)} an intractable $O(N^3)$ computation to compute it exactly, and \textbf{2)} storing the full training dataset, which breaks the requirements of continual learning. Sparse variational inducing variable methods solve these problems \citep{Titsias2009b}, by introducing a small number of $M$ inducing points that control the capacity of the posterior approximation. In certain settings (e.g.~bounded inputs), this approximation is near-exact even when $M \ll N$ \citep{burt2019rates,Burt2020}. This property has allowed continual learning methods to be developed for GPs that perform very closely to full-batch methods \citep{Bui2017,Maddox,Chang2023}, \emph{provided} $M$ is large enough.

As in neural network models, automatically selecting the capacity $M$ is an open problem, with several proposed solutions. \citet{kapoor2021variational} acknowledge the need for scaling the capacity with data size, and propose VAR-GP (Variational Autoregressive GP) which adds a fixed number of inducing points for every batch. However, this number may be too small, leading to poor performance, or too large, leading to wasted computation. \citet{Galy-Fajou2021-tu} propose OIPS (online inducing point selection), which determines $M$ through a threshold on the correlation with other inducing points, and is currently the only adaptive method for GPs, although we show that it still relies on tuned hyperparameters to work.

In this work, we propose to select the capacity of the variational approximation by selecting an appropriate tolerance in the KL gap to the true posterior. This criterion works within the same computational constraints as existing GPs continual learning methods, adapts the capacity to the dataset to minimise computational waste while retaining near-optimal predictive performance. Our method has a single hyperparameter that we keep fixed to a single value, which produces similar trade-offs between size and performance across benchmark datasets with different characteristics.

\section{Background}
\subsection{Sparse Variational Gaussian Processes}
We consider the typical regression setting, with training data consisting of $N$ input/output pairs $\{\mathbf{x}_n, y_n\}_{n=1}^N, \mathbf{x}_n \in \mathbb{R}^D, y_n \in \mathbb{R}$. We model $y_n$ by passing $\mathbf{x}_n$ through a function followed by additive Gaussian noise $y_n = f(\mathbf{x}_n) + \epsilon_n, \epsilon_n \sim \mathcal{N}(0, \sigma^2)$, and take a Gaussian process prior on $f \sim \mathcal{GP}(0, k_\theta(\cdot,\cdot))$ with zero mean, and a kernel $k$ with hyperparameters $\theta$. We wish to find the posterior (for prediction) and marginal likelihood (for finding $\theta$). We need approximations, despite these quantities being available in closed form \citep{RasmussenWilliams2005}, since they have a computational cost of $O(N^3)$ that is too high, and require all training data (or statistics greater in size) to be stored, both of which are prohibitive for continual learning. Variational inference requires a lower $O(NM^2)$ computational and $O(NM)$ memory cost by selecting an approximation from a set of tractable posteriors:
\begin{align}
 q(f(\cdot)) &= \int p(f(\cdot)|\mathbf u,\theta) q(\mathbf u) \mathrm d \mathbf u \label{eq:q-construction} \\
    &= \mathcal{N}\left(f(\cdot); \kcu\kuu^{-1}\mathbf{m},  \right. \nonumber \\
    &\quad \left. \kcc - \kcu\kuu^{-1} (\kuu - \mathbf{S}) \kuu^{-1}\kuc\right) \,, \label{eq:q-eqn}
\end{align}
with $[\kuu]_{ij} = k(\mathbf z_i, \mathbf z_j)$, $[\kcu]_i = [\kuc^T]_i = k(\cdot, \mathbf z_i), \mathbf Z = \{\mathbf z_m\}_{m=1}^M$, and $q(\mathbf u) = \mathcal{N}(\mathbf u; \mathbf m, \mathbf S)$. The variational parameters $\mathbf m, \mathbf S, \mathbf Z$ and hyperparameters $\theta$ are selected by maximising the Evidence Lower Bound (ELBO) \citep{Hensman2013}. This simultaneously minimises the KL gap $\mathrm{KL}[q(f)\,||\,p(f |\mathbf{y}, \theta)]$ between the approximate and true GP posteriors \citep{matthews2016sparse,matthews2017thesis}, and maximises an approximation to the marginal likelihood of the hyperparameters:
\begin{multline}
\mathcal{L}_\text{ELBO} = \sum_{i = 1}^N \mathbb{E}_{q(f(\mathbf x_i))}[\log p(y_i|f(\mathbf x_i), \theta)] \\ - \mathrm{KL}\left[q(\mathbf{u}) \,\|\, p(\mathbf{u} | \theta )\right]\;.\label{eq:elbo}
\end{multline}
The variational approximation has the desirable properties \citep{vdw2019sparse} of \textbf{1)} providing a measure of discrepancy between the finite capacity approximation, and the true infinite capacity model, \textbf{2)} arbitrarily accurate approximations if enough capacity is added \citep{burt2019rates,Burt2020}, and \textbf{3)} retaining the uncertainty quantification over the infinite number of basis functions. In this work, we will particularly rely on being able to measure the quality of the approximation to help determine how large $M$ should be.

\subsection{Sparse Gaussian Processes are Equivalent to Single-Layer Neural Networks}
The equations of the predictor (eq.~\eqref{eq:q-eqn}) show the strong connection between Sparse Gaussian processes and neural networks. The kernel $k(\cdot, z_i)$ forms the nonlinearity, with the weights being parameterised as $\kuu^{-1}\mathbf m$. For inner product kernels $k(\mathbf x, \mathbf Z) = \sigma(\mathbf Z \mathbf x)$ like the arc-cosine kernel \citep{cho2009kernel}, the connection becomes stronger, with $\mathbf Z$ as the input weights. This construction also arises from other combinations of kernels and inter-domain inducing variables \citep{dutordoir2020sparse,sun2021neural}, and has also showed equivalences between deep Gaussian processes and deep neural networks with activations similar to the ReLU \citep{dutordoir2021deep}. As a consequence, our method for determining the number of inducing variables in a sparse GP, equivalently finds the number of neurons needed in a single-layer neural network. As such, we hope the ideas presented in this work can inspire similar mechanisms for adaptive neural networks.

\subsection{Online Sparse Gaussian Processes}
In this work, we use the extension of the sparse variational GP approximation to the continual learning case developed by \citet{Bui2017}. Here, we present a modified derivation that clarifies \textbf{1)} how the online ELBO provides an estimate to the full-batch ELBO, and \textbf{2)} when this approximation is accurate.

We update our posterior and hyperparameter approximations after each batch of new data $\{\mathbf{X}_{n}, \mathbf{y}_{n}\}$. While we do not have access to data from older batches $\{\mathbf{X}_{o}, \mathbf{y}_{o}\}$, the parameters specifying the approximate posterior $q_{o}(f) =  p(f_{\neq \mathbf{a}}| \mathbf{a}, \theta_{o})q_{o}(\mathbf{a})$ are passed on. We construct this approximate posterior as in eq.~\eqref{eq:q-construction} but with $\mathbf{a} = f(\mathbf{Z}_{o})$ and the old hyperparameters $\theta_o$. 
Given the ``old'' $q_o(f)$, online sparse GPs construct a ``new'' approximation $q_{n}(f)$ of the posterior for all observed data $p(f|\mathbf{y}_{o},\mathbf{y}_{n}, \theta_{n})$, which can be written as:
\begin{equation*} 
\begin{aligned}
p(f|\mathbf{y}_{o},\mathbf{y}_{n}, \theta_{n}) &= \frac{p(f|\theta_{n})p(\mathbf{y}_{n}|f)p(\mathbf{y}_{o}|f)}{p(\mathbf{y}_{n}, \mathbf{y}_{o}|\theta_{n})}
\\ &= \frac{p(f|\theta_{n})p(\mathbf{y}_{n}|f)}{p(\mathbf{y}_{n}, \mathbf{y}_{o}|\theta_{n})}\frac{p(f|\mathbf{y}_{o}, \theta_{o})p(\mathbf{y}_{o}|\theta_{o})}{p(f|\theta_{o})}\;.
\end{aligned}
\end{equation*}
We denote the new variational distribution as $q_{n}(f)=  p(f_{\neq \mathbf{b}}| \mathbf{b}, \theta_{n})q_{n}(\mathbf{b})$ where $\mathbf{b} = f(\mathbf{Z}_{n})$ and $\theta_n$ is the new hyperparameter setting.
The KL divergence between the exact and approximate posterior at the current batch is given by: 
\begin{align*}\label{KL:online}
&\operatorname{KL}[q_{n}(f)\mid\mid p(f|\mathbf{y}_{o},\mathbf{y}_{n}, \theta_{n})] 
\\ &
=\log\frac{p(\mathbf{y}_{n}, \mathbf{y}_{o}|\theta_{n})}{p(\mathbf{y}_{o}|\theta_{o})} \\ &\qquad -  \int  q_{n}(f)\log \frac{ 
p(f|\theta_{n})p(\mathbf{y}_{n}|f)p(f|\mathbf{y}_{o}, \theta_{o})}{ q_{n}(f)p(f|\theta_{o})}\text{d}f\;.
\end{align*}
The posterior distribution $p(f|\mathbf{y}_{o}, \theta_{o})$ is not available, however by multiplying its approximation $q_o(f)$ in both sides of the fraction inside the log, we obtain: 
\begin{equation}
\begin{aligned}
&\operatorname{KL}[q_{n}(f)\mid\mid p(f|\mathbf{y}_{o},\mathbf{y}_{n}, \theta_{n})]\\ &=\log\frac{p(\mathbf{y}_{n}, \mathbf{y}_{o}|\theta_{n})}{p(\mathbf{y}_{o}|\theta_{o})} 
\\& \quad -
\underbrace{\int  q_{n}(f)\log \frac{ p(f|\theta_{n})p(\mathbf{y}_{n}|f)q_{o}(f)}{q_{n}(f)p(f|\theta_{o})}\text{d}f}_{=\hat{\mathcal{L}}} + \Phi\label{eq:KLDhatL}
\end{aligned}
\end{equation}
where $\Phi = -\int q_{n}(f) \log  \frac{p(f|\mathbf{y}_{o}, \theta_{o})}{q_{o}(f)}\text{d}f$. We cannot compute $\Phi$ due to its dependence on the exact posterior, so we drop it and use the remaining term as our ``online ELBO'' training objective, which simplifies as:
\begin{equation}
\widehat{\mathcal{L}} :=   \int q_{n}(f)\left[\log \frac{p(\mathbf{b}|\theta_{n})q_{o}(\mathbf{a})p(\mathbf{y}_{n}|f)} {q_{n}(\mathbf{b})p(\mathbf{a}|\theta_{o})}\right]\text{d}f, 
\label{hatL}
\end{equation}
since  $q_o(f)=p\left(f_{\neq \mathbf{a}} \mid \mathbf{a}, \theta_o\right) q_o(\mathbf{a})$ and similarly for $q_n(f)$. 

Maximising $\widehat{\mathcal{L}}$ will accurately minimise the KL to the true posterior when $\Phi$ is small, which is the case when the old approximation is accurate, i.e.~$q_o(f) \approx p(f|\mathbf{y}_{o}, \theta_{o})$ for all values of $f(X_o)$ (with $\Phi = 0$ in the case of equality). In our continual learning procedure, we will keep our sequence of approximations accurate by ensuring they all have enough inducing points, which will keep $\Phi$ small.

To get our final bound, we perform a change of variables for the variational distribution $q_o(\mathbf{a}) = \mathcal{N}(\mathbf{a}; \mathbf{m}_\mathbf{a}, \mathbf{S}_\mathbf{a})$ to use the likelihood parametrisation \citep{panos2018fully}:
\begin{equation}
\begin{aligned}
q_{o}(\mathbf{a}) &=  \frac{\mathcal{N}(\mathbf{a}; \tilde{\mathbf{m}}_\mathbf{a}, \mathbf{D}_\mathbf{a}) \mathcal{N}(\mathbf{a}; 0, \mathbf{K}^\prime_{\mathbf{aa}})}{\int \mathcal{N}(\mathbf{a}; \tilde{\mathbf{m}}_\mathbf{a}, \mathbf{D}_\mathbf{a}) \mathcal{N}(\mathbf{a}; 0, \mathbf{K}^\prime_{\mathbf{aa}}) \text{d}\mathbf{a}} \\
&= \frac{l(\mathbf{a})p(\mathbf{a} \mid  \theta_{o})}{\mathcal{N}(\mathbf{a}; 0, \mathbf{D}_\mathbf{a} + \mathbf{K}^\prime_{\mathbf{aa}})}, 
\end{aligned}
\end{equation}
where $\mathbf{D}_{\mathbf{a}}=\left(\mathbf{S}_{\mathbf{a}}^{-1} - \mathbf{K}_{\mathbf{a a}}^{\prime-1}\right)^{-1}$ and $\tilde{\mathbf{m}}_\mathbf{a}=\mathbf{K}_{\mathbf{aa}}^{\prime -1}\mathbf{m}_\mathbf{a} $ are the variational parameters, $\mathbf{K}_{\mathbf{aa}}^\prime$ is the covariance for the prior distribution $p(\mathbf{a} \mid \theta_{o})$ and $l(\mathbf{a}) := \mathcal{N}(\mathbf{a}; \tilde{\mathbf{m}}_\mathbf{a}, \mathbf{D}_\mathbf{a})$. In this formulation, the variational parameters $\tilde{\mathbf{m}}_{\bf a}, \mathbf{D}_{\bf a}$ effectively form a dataset that produce the same posterior as the original dataset, but which we have chosen to be smaller in size, $M < N$. This makes our online ELBO from eq.~\eqref{hatL}
\begin{multline}\label{eq:Lhatlikelihood}
\widehat{\mathcal{L}} = \mathbb{E}_{q_{n}(f)}\left[\log p(\mathbf{y}_{n} | f) \right] +  \mathbb{E}_{q_{n}(f)}\left[\log l(\mathbf{a})\right] \\
- \operatorname{KL}\left[q_{n}(\mathbf{b}) \mid \mid p(\mathbf{b} | \theta_{n})  \right]
 - \log \mathcal{N}(\mathbf{a}; 0,   \mathbf{K}^\prime_{\mathbf{aa}} + \mathbf{D}_\mathbf{a})\,,
\end{multline}
which has the nice interpretation of being the normal ELBO, but with an additional term that includes the approximate likelihood $l(\mathbf a)$ which summarises the effect of all previous data.

While $\widehat{\mathcal{L}}$ is all that is needed to train the online approximation, it differs from the true marginal likelihood by the term $\log p(\mathbf y_o | \theta_o)$. To approximate it, we could drop the term $\log \mathcal N(\mathbf a; 0, \mathbf K'_{\bf aa} + \mathbf D_{\bf a})$ from $\widehat{\mathcal{L}}$, since this term also approximates $\log p(\mathbf y_o|\theta_o)$, with equality when the posterior is exact, but with no guarantee of being a lower bound.

Although $\widehat{\mathcal{L}}$ is a useful training objective for general likelihoods, the regression case we consider allows us to analytically find $q(\mathbf b)$ \citep{Bui2017} (see App.\,\ref{appendix:lowerbound}), resulting in the lower bound
\begin{multline}
    \label{eq:onlinecollapsed}
    \widehat{\mathcal{L}}= \log \mathcal{N}\left(\hat{\mathbf{y}} ; \mathbf{0}, \mathbf{K}_{\hat{\mathbf{f}} \mathbf{b}} \mathbf{K}_{\mathbf{b b}}^{-1} \mathbf{K}_{\mathbf{b} \hat{\mathbf{f}}}+\Sigma_{\hat{\mathbf{y}}}\right)  + \Delta \\
    - \frac{1}{2} \operatorname{tr}\left[\mathbf{D}_\mathbf{a}^{-1}(\mathbf{K}_\mathbf{aa} - \mathbf{Q}_\mathbf{aa} )  \right]- \frac{1}{2\sigma^2} \operatorname{tr}(\mathbf{K}_\mathbf{ff} - \mathbf{Q}_\mathbf{ff} ) 
\end{multline}
where $\mathbf{K}_{\hat{\mathbf{f}} \mathbf{b}} = \left[
 \mathbf{K}_{\mathbf{fb}} \quad \mathbf{K}_{\mathbf{ab}}\right]^\top$, $\Sigma_{\hat{\mathbf{y}}} = \text{diag}([\sigma_y^2 \mathbf{I},\, \mathbf{D}_{\mathbf{a}}])$, 
 $\hat{\mathbf{y}}=\left[\mathbf{y}_{n}\quad \mathbf{D}_{\mathbf{a}} \mathbf{S}{\mathbf{a}}^{-1} \mathbf{m}_{\mathbf{a}}
\right]^\top$, 
and
\begin{multline*}
    \Delta = -\frac{1}{2} \log \frac{|\mathbf{S}_\mathbf{a}|}{|\mathbf{K}_\mathbf{aa}^\prime||\mathbf{D}_\mathbf{a}|} + \frac{M_\mathbf{a}}{2} \log (2\pi) \\-\frac{1}{2}\mathbf{m}_\mathbf{a}^T\mathbf{S}_\mathbf{a}^{-1}\mathbf{m}_\mathbf{a} + \frac{1}{2}\mathbf{m}_\mathbf{a}^T\mathbf{S}_\mathbf{a}^{-1}\mathbf{D}_\mathbf{a}\mathbf{S}_\mathbf{a}^{-1}\mathbf{m}_\mathbf{a}, 
\end{multline*}
with $\mathbf{Q}_\mathbf{ff} = \mathbf{K}_\mathbf{fb}\mathbf{K}_\mathbf{bb}^{-1}\mathbf{K}_\mathbf{bf} $, $\mathbf{Q}_\mathbf{aa} = \mathbf{K}_\mathbf{ab}\mathbf{K}_\mathbf{bb}^{-1}\mathbf{K}_\mathbf{ba}$. 
All covariances are computed using the new hyperparameters $\theta_{n}$, except for $\mathbf{K}_{\mathbf{aa}}^\prime$ which is the covariance for the prior distribution $p( \mathbf{a} \mid \theta_{o})$. Finally, $M_{\mathbf{a}}=|\mathbf{a}|$ is the number of inducing points used at the previous batch. For calculating $\widehat{\mathcal{L}}$ in each batch, the computational complexity is $O(N_{n}M_\mathbf{b}^2 + M_\mathbf{b}^3)$ and the memory requirements are $O(M_\mathbf{b}^2)$ where $M_\mathbf{b}$ is the total number of inducing points for the current batch.  

To create a fully black-box solution, we still need to specify how to select the hyperparameters $\theta_n$, the number of inducing variables $M_\mathbf{b}$, and the inducing inputs $\mathbf Z_n$. We will always select $\theta_n$ by maximising $\widehat{\mathcal{L}}$ using L-BFGS. To select the locations $\mathbf Z_n$, we use the ``greedy variance'' criterion \citep{fine2001efficient, foster2009stable, Burt2020}. This leaves only the number of inducing variables $M_\mathbf{b}$ to be selected.

\section{Automatically Adapting Approximation Capacity}
We propose a method for adjusting the capacity of the approximation $M_{\bf b}$ to maintain accuracy. We propose to keep inducing points from old batches fixed, and select new inducing points from each incoming batch, with their locations set using the ``greedy variance'' criterion \citep{fine2001efficient, foster2009stable, Burt2020}. While optimising all inducing points does lead to a strictly better approximation, we avoid this for the sake of simplicity. The question remains: To achieve a level of accuracy, ``how big is big enough?'' To answer this, we consider the online ELBO as a function of the capacity $\widehat{\mathcal{L}}(M_{\bf b})$, and propose a threshold after which to stop adding new inducing variables.

\subsection{Online Log Marginal Likelihood (LML) Upper Bound}
The problem of selecting a sufficient number of inducing variables is also still open in the batch setting. One possible strategy is to use an upper bound to the marginal likelihood \citep{titsias2014} to bound $\mathrm{KL}[q(f)||p(f|\mathbf y)] \leq \mathcal U - \mathcal L$, and stop adding inducing variables once this is below a tolerance $\alpha$. To extend this strategy to online learning, we begin by deriving an online upper bound, as a counterpart to the online ELBO from eq.~\eqref{eq:onlinecollapsed}. We follow the same strategy as \citet{titsias2014}, by considering the highest possible value that our lower bound can attain. While in full-batch inference this is equal to the true LML, in our case this is obtained by keeping the inducing inputs from the previous iteration, and adding each new datapoint to the inducing set:
\begin{equation}\label{eq:Lstar}
    \mathcal{L}^* := \widehat{\mathcal{L}}(N_n+M_{\bf a}) = \log \mathcal{N} \left(
             \hat{\mathbf{y}} ;\,\mathbf{0}\,,  \mathbf{K}_{\hat{\mathbf{f}}\hat{\mathbf{f}}} + \Sigma_{\hat{y}}\right)  +  \Delta_\mathbf{a}
\end{equation}
with $\quad \mathbf{K}_{\hat{\mathbf{f}}\hat{\mathbf{f}}} = \left[\begin{array}{cc}
                \mathbf{K}_{\mathbf{ff}} & \mathbf{K}_{\mathbf{fa}} \\
                \mathbf{K}_{\mathbf{af}} & \mathbf{K}_{\mathbf{aa}}
                \end{array}\right].$
Using properties of positive semi-definite matrices, we derive an upper bound $\widehat{\mathcal{U}}(M)$ to eq. \eqref{eq:Lstar}:
\begin{equation}
\begin{aligned}
 & \log \mathcal{N} \left(
            \hat{\mathbf{y}} ;\,\mathbf{0}\,,  \mathbf{K}_{\mathbf{\hat{f}\hat{f}}} + \Sigma_{\hat{y}}\right)\\
&\leq -\frac{(N+ M_\mathbf{a})}{2} \log(2\pi) \nonumber 
 - \frac{1}{2} \log  |\mathbf{K}_{\mathbf{\hat{f}b}}\mathbf{K}_{\mathbf{bb}}^{-1}\mathbf{K}_{\mathbf{b\hat{f}}} + \Sigma_{\mathbf{\hat{y}}} |  \nonumber \\
&\quad - \frac{1}{2} \hat{\mathbf{y}}^T\left( \mathbf{K}_{\mathbf{\hat{f}b}}\mathbf{K}_{\mathbf{bb}}^{-1}\mathbf{K}_{\mathbf{b\hat{f}}} +t\mathbf+ \Sigma_{\mathbf{\hat{y}}} \right)^{-1}\hat{\mathbf{y}}  \nonumber \\ 
& \coloneqq \mathcal{\widehat{U}}(M),
\end{aligned}
\end{equation}
where $t = \operatorname{tr}(\mathbf{K}_{\mathbf{\hat{f}\hat{f}}} - \mathbf{Q}_{\mathbf{\hat{f}\hat{f}}})$ and 
$\mathbf{Q}_{\mathbf{\hat{f}\hat{f}}} = \mathbf{K}_{\mathbf{\hat{f}b}}\mathbf{K}_{\mathbf{bb}}^{-1}\mathbf{K}_{\mathbf{b\hat{f}}}$ and $M$ is the number of inducing points used to calculate the bound (which can be unequal to $M_\mathbf{b}$).

\begin{algorithm*}[t]
\caption{Vegas Inducing Point Selection (VIPS)}\label{alg:main}
\begin{algorithmic}
   \STATE {\bfseries Input:} $\mathbf{X}_{n} = \{\mathbf{x}_i\}_{i=1}^{N_{n}}$, $\mathbf{Z}_{o} =  \{\mathbf{z}_m\}_{m=1}^{M_{\mathbf{a}}}$, $\hat{\mu}$, $\hat\sigma$, $\theta_{n}$, kernel $k(\cdot, \cdot | \theta_{n})$, threshold parameter $\delta$. 
   \STATE {\bfseries Output:} Updated set \(\mathbf{Z}_{n} = \mathbf{Z}_{o} \cup \{\mathbf{x}_{m'}\}_{m'=1}^{M'}\), where \(|\mathbf{Z}_{n}| = M_\mathbf{b}\).
   \STATE Initialise $\mathbf{Z}_{n} =\mathbf{Z}_{o}$. 
   \WHILE{$\widehat{\mathcal{U}}(M)-\widehat{\mathcal{L}}(M_{\mathbf{b}}) \leq \delta|\widehat{\mathcal{U}}(M) - \mathcal{L}_{noise}(\hat{\mu}, \hat{\sigma})|$}
    \STATE Select $ \mathbf{x} = {\operatorname{argmax}}_{ \mathbf{x} \in \mathbf{X}_n}\, k(\mathbf{x}, \mathbf{x}) - \mathbf{k}_\mathbf{b}(\mathbf{x})^\top\mathbf{K}_\mathbf{bb}^{-1} \mathbf{k}_\mathbf{b}(\mathbf{x})$. 
   \STATE Add  $\mathbf{x}$ to the set of inducing points: $\mathbf{Z}_{n} = \mathbf{Z}_{n}\cup\{\mathbf{x}\}$.
   \ENDWHILE
\end{algorithmic}
\end{algorithm*}

\subsection{Approximation Quality Guarantees}\label{sec:approx}

Adding inducing points will eventually increase $\widehat{\mathcal{L}}$ until it reaches $\mathcal{L}^*$ \citep{bauer2016,matthews2017thesis,Burt2020}. If we add inducing points until $\widehat{\mathcal{U}}(M) - \widehat{\mathcal{L}}(M_\mathbf{b}) \leq \alpha$ we can guarantee the following:
\begin{guarantee}\label{thm:close}
Let $M$ be a fixed integer and $M_\mathbf{b}$ be the number of selected inducing points such that $\widehat{\mathcal{U}}(M) - \widehat{\mathcal{L}}(M_\mathbf{b}) \leq \alpha$. Assuming that $\theta_n=\theta_o$, we have two equivalent bounds:
\begin{align}
\operatorname{KL}[q_{n}(f)\mid\mid p(f|\mathbf{y}_{o}, \mathbf{y}_{n}, \theta_{o})]&\leq \alpha + \Psi   \\
\mathrm{KL}[q_n(f)\mid\mid q_n^*(f)] &\leq \alpha
\end{align}
where $\Psi = \int q_n(f) \log \frac{q_n^*(f)}{p(f|\mathbf y_n, \mathbf y_o)} \mathrm d f$ and $q_n^*(f) = \mathcal{Z}^{-1} q_o(f) p(\mathbf{y}_n \mid f)$ represents the variational distribution associated with the optimal lower bound $\mathcal{L}^* = \widehat{\mathcal{L}}(N_n + M_{\mathbf{a}})$, with $\mathcal{Z}$ denoting the marginal likelihood that normalises $q_n^*(f)$.
\end{guarantee}
\begin{proof}
We cease the addition of points when $\widehat{\mathcal{L}}(M_\mathbf{b}) > \widehat{\mathcal{U}}(M) - \alpha $. Given that $\widehat{\mathcal{U}}(M) \geq \mathcal{L}^*$, and assuming $\theta_n = \theta_o$, the rest follows from algebraic manipulation of eq.~\eqref{eq:KLDhatL}. See App.\,\ref{appendix:proof} for the complete proof. 
\end{proof}
The first bound shows that if $\Psi$ is near zero, the KL to the true posterior is bounded by $\alpha$. While $\Psi$ depends on the true posterior and therefore cannot be computed, if the posterior in the previous iteration was exact, $\Psi$ would be equal to zero. The second bound shows that we are guaranteed to have our actual approximation $q_n(f)$ be within $\alpha$ nats of the best approximation that we can develop, given the limitations of the approximations made in previous iterations.

\subsection{Selecting a Threshold}
In this final step of our online learning method, we must specify a heuristic for selecting $\alpha$ that does not require knowing any data in advance, while also working in a uniform way across datasets with different properties. A constant value for $\alpha$ does not work well, since the scale of the LML depends strongly on properties such as dataset size, and observation noise. This means that a tolerance of 1 nat \citep{cover1999elementsnats} may be appropriate for a small dataset, but not for a large one. 

As a principle for selecting the threshold, we take loose inspiration from compression and MDL \citep{grunwald2019minimum}, which takes the view of the ELBO being proportional to negative the code length that the model requires to encode the dataset. Intuitively, our desire to select an $\alpha$ such that our method captures a high proportion (e.g.~95\%) of all the information in each batch, so that we can compress to within a small fraction of the optimal variational code. By itself, this is not well-defined since the code length depends on the quantisation tolerance for our continuous variables. To avoid this, we take an independent random noise code as our baseline, and select $\alpha$ to be within some small fraction of the optimal variational code, relative to the random noise code. We want  to capture a high proportion of all the additional information that our model provides, relative to the noise model, i.e.~we want our threshold to be:
\begin{align*}
\alpha = \delta (\mathcal{L}^* - \mathcal{L}_{\text{noise}})\,,\quad \mathcal{L}_{\text{noise}} = \sum_{n=1}^{N_n} \log \mathcal{N}(y_n; \hat{\mu}, \hat{\sigma}^2)
\end{align*}
where $\hat{\mu}$ and $\hat{\sigma}^2$ are the average and variance of the observations for up to the current task and $\delta$ is a user-defined hyperparameter. We validate that this approach leads to values of $\delta$ giving consistent behaviour across a wide range of datasets, which allows it to be set in advance without needing much prior knowledge of the dataset characteristics.

Calculating this threshold is intractable for large batch sizes $N_n$. However, if we change our stopping criterion to the more stringent upper bound
\begin{align}
    \bar{\alpha} = \delta(\widehat{\mathcal{U}}(M) - \mathcal{L}_{\text{noise}})\,.
\end{align}
and increase $M$ for calculating $\widehat{\mathcal{U}}$ as $M_\mathbf{b}$ is increased for calculating $\widehat{\mathcal{L}}$, we obtain the same guarantees as before but at a lower computational cost. However, this strategy is only worthwhile for very large batch sizes $N_n$, due to the importance of constant factors in the computational cost. In the common continual learning settings we investigate $N_n$ is small enough to allow computing $\mathcal{L}^*$.

The procedure for our inducing point selection method is detailed in \cref{alg:main} and we give further details in App.~\ref{append:algo}. We name our approach Vegas Inducing Point Selection (VIPS), drawing an analogy to Las Vegas Algorithms. These methods guarantee the accuracy of the output, however, their computational time fluctuates for every run  \citep{VegasBook}.

\section{Experiments}
We evaluate the performance of our adaptive inducing point selection method VIPS in a range of streaming scenarios where we assume the total number of observations is unknown. In all cases, the variational distribution, noise and kernel hyperparameters are optimised using the online lower bound (Eq. \eqref{eq:onlinecollapsed}). 
We provide further details in App.\,\ref{appendix:experiments}.

Continual learning scenarios pose unique challenges. It is impossible to pre-determine memory allocation due to unknown input space coverage. Additionally, cross-validation for hyperparameter tuning is not feasible as it would require storing all data. Thus, an effective method must \textbf{1)} have an adaptive memory that can grow with the demands of the data, \textbf{2)} work with hyperparameters that can be set before training. The first part of our experiments demonstrates the benefits of using an adaptive model capacity in continual learning. The second part highlights the necessity of a single, pre-tunable hyperparameter that works across diverse settings. We benchmark VIPS against two other inducing point selection methods on streaming UCI datasets and a real-world robotics application.

\subsection{Model size and data distribution}
Figure \ref{fig:data_types} shows VIPS's ability to adapt across three datasets with different characteristics, each divided into ten batches, illustrating how input distribution drives model growth as more data is seen. In the first dataset, each batch introduces data from new parts of the input space. 
Since each batch is equally novel, the model size grows linearly. In the second, the data remain within a fixed interval, leading to diminishing novelty in each batch, and a model size that converges to a fixed value. The third dataset mixes narrow-range uniform samples with occasional batches sampled from a heavy-tailed Cauchy distribution. This leads the model size to converge, with sporadic growth when novel data is observed.

\begin{figure}[t]
\vskip 0.2in
\centering
\begin{subfigure}{.499\textwidth}
\includegraphics[width=0.95\linewidth]{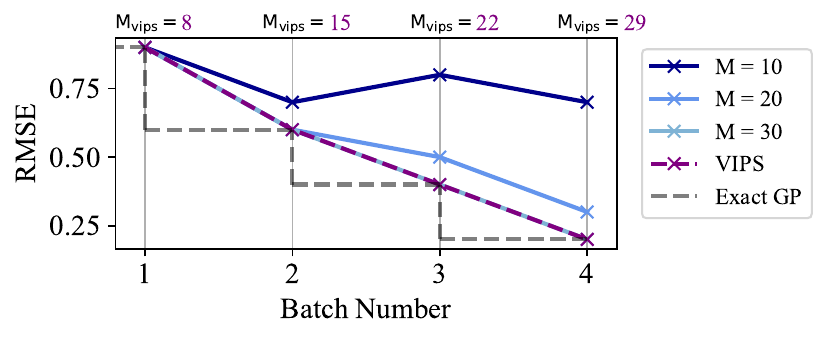} 
\caption{Accuracy on a synthetic dataset.}
\label{fig:toyrmse}
\end{subfigure}
\begin{subfigure}{.499\textwidth}
\includegraphics[width=\linewidth]{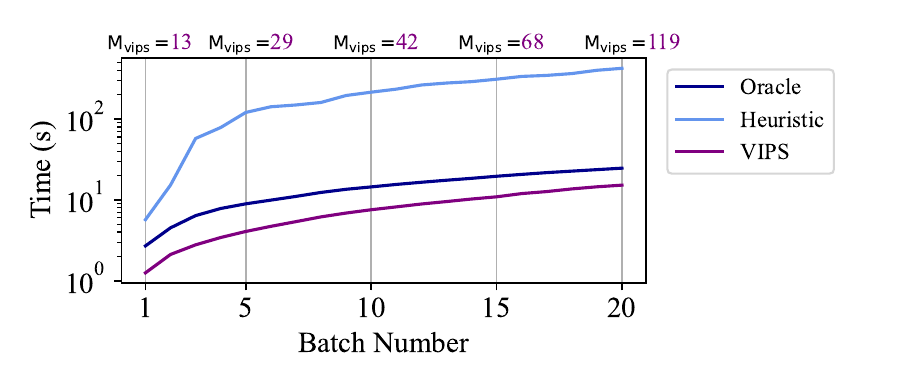}
\caption{Training time comparison on ``naval'' dataset. }
\label{fig:toytime}
 \end{subfigure}
\caption{\textbf{(a)} Performance comparison of fixed memory approaches (blue curves with  $M = 10, 20, 30$ inducing points) and VIPS, with $M$ (shown at the top) inducing points at each batch.
\textbf{(b)} (log) Time taken to train the online GP model on the ``naval'' dataset divided into 20 batches with fixed size (oracle: $M=100$, heuristic: $M=1000$) and VIPS, our adaptive method. }
\label{fig:toymain}
\vskip -0.2in
\end{figure}

\subsection{The impact of model capacity on accuracy and training time}
Here, we compare a fixed to dynamic model size, with inducing points selected according to the greedy variance criterion \citep{Burt2020}, to demonstrate the challenges of a fixed model size. Since the total dataset size and data characteristics are not known at the start of training, selecting a fixed size will either lead to a model that is too small and underperforms, or a model that is too large and wastes computational resources. Adaptive size solves these issues.

\textbf{Performance comparison}: We train on a synthetic dataset divided into four batches, using a fixed model size, and our adaptive stopping criterion, VIPS. We test three fixed sizes: $M = 10,\, 20$, and $30$ inducing points. We record the test root mean square error (RMSE) and compare to an exact GP model with access to all current training data as a benchmark. Figure \ref{fig:toyrmse} shows that fixed size models $M = 10, 20$ lose accuracy with more data whereas fixed model $M\!=\!30$ and VIPS match exact GP performance. At this point, we can select the model with $M=30$ with the benefit of hindsight. However, at the start of training, we could not have known that this size would have been sufficient at the point of testing. VIPS not only automatically ends up with this optimal size, but also avoids computational waste by using fewer inducing points in earlier batches.

\textbf{Training cost comparison}: We use the ``naval'' UCI dataset divided into 20 batches, to compare the training time of models with a fixed and dynamic size. We compare a best-case fixed-size model with the smallest number of inducing points that would still give near-optimal performance. For ``naval'', this is $M=100$, which would have to be set by an ``oracle''. We also include a typical heuristic, of $M = 1000$ (around 1/10th of the total data points), that would ensure sufficient capacity to handle different data patterns and complexities. 
Finally, we test our adaptive method, VIPS, which adjusts the number of inducing points as needed while achieving near-exact performance for the dataset. As shown in Figure \ref{fig:toytime}, VIPS results in lower computational time by only growing its model size as new data is seen, beating even the oracle method.

\subsection{Continual learning of UCI datasets}
\label{sec:uci}
 We compare our method, VIPS, to two other inducing point selection approaches: Conditional Variance (CV) and OIPS \citep{Galy-Fajou2021-tu} (see details in App.~\ref{appendix:methods}). We use six datasets from the UCI repository \citep{UCI}, simulating a continual learning scenario by sorting the data along the first dimension and dividing it into batches. When evaluating inducing point selection methods, we must consider the trade-off between accuracy and model size: while selecting more inducing points improves model accuracy, it also increases computational cost. 

\begin{table}[t]
\caption{Mean (std) of inducing points in final batch across training/test splits at operating point selected to achieve 10\% RMSE threshold. Cross (\xmark) indicates failure to meet accuracy threshold and "Max." denotes reaching maximum capacity limit. Values for OIPS are in italics, as the operating point was selected based on only five datasets, making it less stringent than other methods.}
\label{tab:comparison}
\vskip 0.15in
\begin{center}
\begin{small}
\begin{sc}
\resizebox{\linewidth}{!}{%
\begin{tabular}{lccc}
\toprule
Dataset & CV & OIPS & VIPS (Ours) 
\\
\midrule
Concrete & 492(54) & \textit{240(102)} & \textbf{234(116)} \\
Skillcraft  &  739(34) &  \textit{174(16)}  & \textbf{134(9)}\\
Kin8nm  &  6316(10) &  \textit{6458(8) } & \textbf{1904(41)}\\
Naval  & \textbf{49(1)} & \xmark   & 57(1)\\
Elevators  & 3133(140) &   \textit{298(8)} &  \textbf{291(4)} \\
Bike  & Max. 7000 &  \textit{1964(137)} & \textbf{650(17)} \\
\bottomrule
\end{tabular}
}
\end{sc}
\end{small}
\end{center}
\vskip -0.1in
\end{table}

\begin{table}[t]
\caption{Mean (std) of inducing points in final batch across training/test splits at operating point selected to achieve 10\% RMSE threshold in all but one dataset. \xmark\,\textcolor{gray}{Gray} shows cases exceeding threshold, with percentages showing \% difference to full-batch GP relative to a noise model.}
\label{tab:comparisonoutlier}
\vskip 0.15in
\begin{center}
\begin{small}
\begin{sc}
\resizebox{\linewidth}{!}{%
\begin{tabular}{lccc}
\toprule
Dataset & CV & OIPS & VIPS (Ours) 
\\
\midrule
Concrete & 184(74) & 240(102) & \textbf{178(95)} \\
Skillcraft  &  226(22) &  174(16) & \textbf{133(0)}\\
Kin8nm  &  5334(27) &  6458(8)  & \textbf{1563(39)}\\
Naval  & \xmark\,\textcolor{gray}{23(8)}-\textcolor{gray}{323\%} & \xmark\,\textcolor{gray}{16(1)}-\textcolor{gray}{315.28\%} & \textbf{48(3)}\\
Elevators  & 836(27) &  298(8) &  \textbf{281(4)} \\
Bike  & 4895(28) &  \textbf{1964(137)} &\xmark\,\color{gray} 549(14)-10.8\% \\
\bottomrule
\end{tabular}
}
\end{sc}
\end{small}
\end{center}
\vskip -0.1in
\end{table}

\begin{table}[t]
\caption{Mean (std) of inducing points in final batch across training/test splits at operating point selected to achieve 10\% NLPD threshold. Cross (\xmark) indicates failure to meet accuracy threshold and "Max." denotes reaching maximum capacity limit. Values for OIPS are in italics, as the operating point was selected based on only five datasets, making it less stringent than other methods.}
\label{tab:comparisonnlpd}
\vskip 0.15in
\begin{center}
\begin{small}
\begin{sc}
\resizebox{\linewidth}{!}{%
\begin{tabular}{lccc}
\toprule
Dataset & CV & OIPS & VIPS (Ours) 
\\
\midrule
Concrete & 492(54) & \textit{383(90)} & 451(77) \\
Skillcraft  &  1091(107) &  \textit{236(38)}  & \textbf{195(18)}\\
Kin8nm  &  6469(7) &  \textit{6523(9)} & \textbf{5065(46)}\\
Naval  & 903(8) & \xmark   & \textbf{509(15)}\\
Elevators  & 3133(140) &   \textit{386(47)} &  \textbf{487(24)} \\
Bike  & Max. 7000 &  \textit{3916(122)} & \textbf{2794(148)} \\
\bottomrule
\end{tabular}
}
\end{sc}
\end{small}
\end{center}
\vskip -0.1in
\end{table}

\begin{table}[ht]
\caption{Mean (std) of inducing points in final batch across training/test splits at operating point selected to achieve 10\% NLPD threshold in all but one dataset. \xmark\,\textcolor{gray}{Gray} shows cases exceeding threshold, with percentages showing \% difference to full-batch GP relative to a noise model.}
\label{tab:comparisonoutliernlpd}
\vskip 0.15in
\begin{center}
\begin{small}
\begin{sc}
\resizebox{\linewidth}{!}{%
\begin{tabular}{lccc}
\toprule
Dataset & CV & OIPS & VIPS (Ours) 
\\
\midrule
Concrete & \textbf{303(78)} & 383(90) & 434(79) \\
Skillcraft  &  295(13) &  236(38) & \textbf{177(13)}\\
Kin8nm  &  5685(26) &  6523(9)  & \textbf{4281(69)}\\
Naval  & \xmark\,\textcolor{gray}{27(6)}-\textcolor{gray}{2406\%} & \xmark\,\textcolor{gray}{20(1)}-\textcolor{gray}{133.87\%} & \xmark\,\color{gray} 373(19)-11.92\%\\
Elevators  & 1099(50) & \textbf{386(47)} &  438(17) \\
Bike  & 5563(36) &  3916(122) &\textbf{2296(138)} \\
\bottomrule
\end{tabular}
}
\end{sc}
\end{small}
\end{center}
\vskip -0.1in
\end{table}

Our goal is to minimise computational cost, while satisfying the constraint that we reach a fixed target accuracy threshold (e.g.~within 10\% RMSE of a full-batch GP). To benchmark our methods, we select a single hyperparameter across datasets, so that our accuracy constraint is always met, while otherwise using the fewest inducing points. We refer to this hyperparameter as the operating point. For this operating point, we then measure the number of inducing points that the method chooses for each dataset, with a lower result being better. If across datasets, there is a large variation in hyperparameter values needed to achieve the accuracy constraint, selecting the most stringent criterion will cause large numbers of inducing points to be used for other datasets. This simulates a realistic setting, where one cannot know what the optimal hyperparameter value is for each dataset beforehand, and penalises methods that are hard to tune, due to large variations in the required hyperparameter values across datasets.

\begin{figure}[t]
\vskip 0.2in
        \begin{subfigure}{.5\textwidth}
        \centering
        \includegraphics[width=0.85\linewidth]{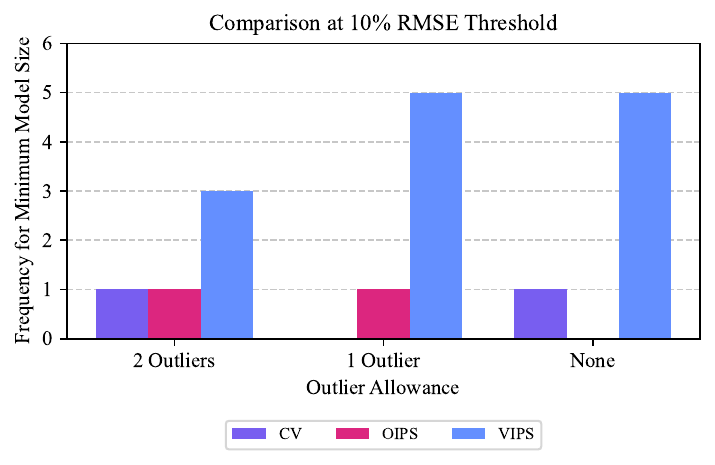} 
    \caption{}
    \label{fig:counts_rmse}
    \end{subfigure}
    \begin{subfigure}{.5\textwidth}
        \centering
        \includegraphics[width=0.85\linewidth]{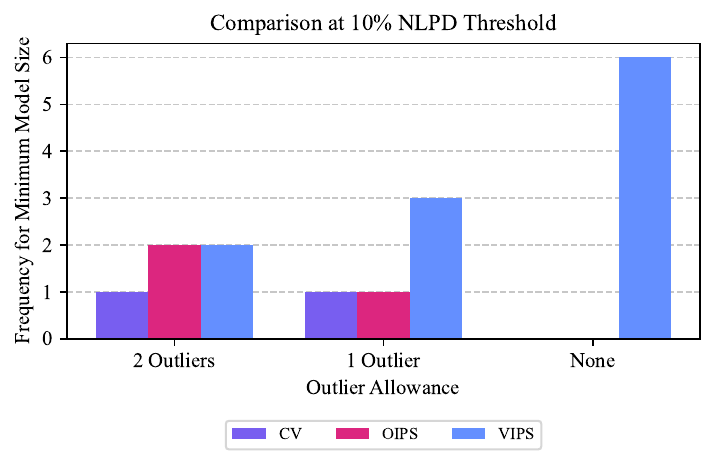} 
\caption{}
\label{fig:counts_nlpd}
\end{subfigure}
\caption{Number of datasets where each method achieves minimal model size at different outlier allowances. A ``win'' is assigned to the dataset with the smallest model which satisfies the \textbf{(a)} RMSE and \textbf{(b)} NLPD thresholds of 10\%. Counts represent absolute wins. Higher counts indicate better method robustness across datasets.
}
\label{fig:counts}
\vskip -0.2in
\end{figure}

Table \ref{tab:comparison} shows the number of inducing points for each method at the operating point that achieves the accuracy threshold within 10\% RMSE of a full-batch GP.  VIPS requires fewer inducing points in the majority of datasets. 
CV often results in larger models, reaching the 7000-point limit on Bike. OIPS performs inconsistently, matching VIPS on smaller datasets but oversizing on larger ones. For the Naval dataset, OIPS failed to meet the accuracy constraint across all tested hyperparameter values. While extending the hyperparameter range might have improved accuracy, it would have led to even larger models. Therefore, we selected an operating point that satisfied our conditions for five out of six datasets, treating Naval as an outlier. Table \ref{tab:comparisonoutlier} shows the results when this exception is applied to all methods. We observe a similar trend, with VIPS achieving the smallest model sizes. 
Tables~\ref{tab:comparisonnlpd} and~\ref{tab:comparisonoutliernlpd} report analogous results for the threshold within 10\% NLPD of a full-batch GP. The same pattern holds. \Cref{fig:counts} summarises these outcomes across outlier allowances, with VIPS more frequently achieving minimal model size. 
 App.\,\ref{appendix:UCI} provides additional results under different thresholds.

Our results show that VIPS performs consistently across datasets, allowing a single pre-set hyperparameter to be used. In contrast, OIPS and CV require tuning, as no single hyperparameter works well for all datasets. This is a limitation in continual learning,  where the whole dataset is unavailable in advance.

\subsection{Continual Learning of Magnetic Field}\label{sec:magneto}
\begin{figure}[t]
\vskip 0.2in
\centering
        \begin{subfigure}{.5\textwidth}
        \centering
        \includegraphics[width=0.6\linewidth]{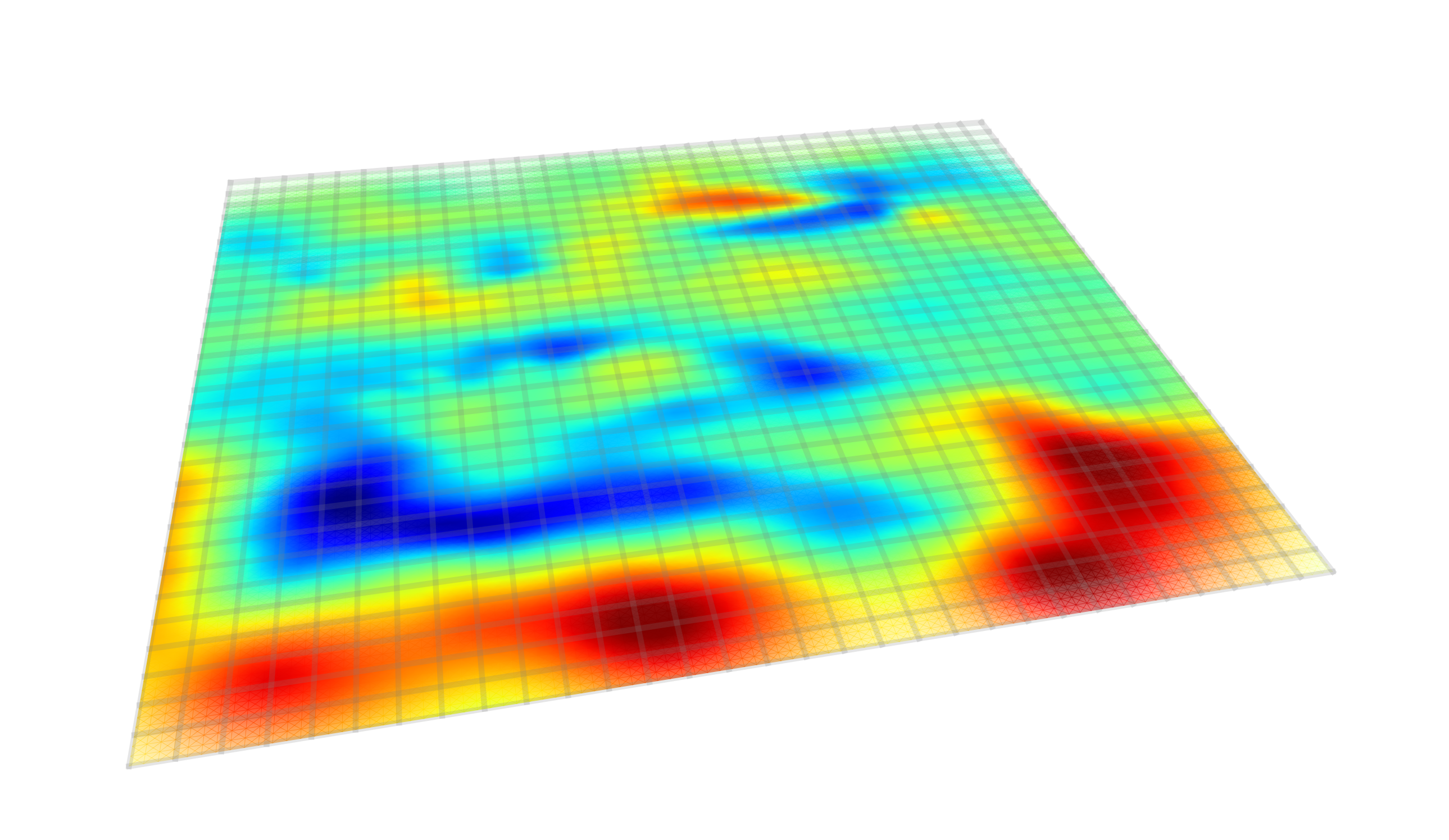} 
    \caption{Learning paths 1, 2, 4, 5 sequentially with VIPS.}
    \label{fig:magneto_2}
    \end{subfigure}
    \begin{subfigure}{.5\textwidth}
        \centering
        \includegraphics[width=0.6\linewidth]{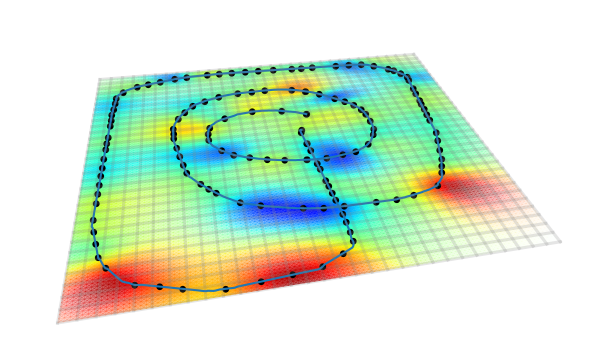}
    \caption{Learning path 3 with VIPS. M: 134, RMSE: 7.55.}
    \label{fig:magneto_1}
    \end{subfigure}
\customcaption{A small robot is used to perform sequential estimation of magnetic field anomalies. The strength of the magnetic field is given by 10 $\mu \mathrm{T}$}{to 90 $\mu \mathrm{T}$. \textbf{(a)} shows the final estimate of the magnitude field learned sequentially through different paths. \textbf{(b)} show the outcome of learning a single path continuously with the black dots representing the chosen inducing points.}
\label{fig:magneto_main}
\vskip -0.2in
\end{figure}

We apply the optimal hyperparameters found in the previous section, selected under the 10\% RMSE threshold, to the experimental setup from \citet{Chang2023} using real-world data from \citet{Solin2018}, where a robot maps magnetic field anomalies in a 6m x 6m indoor space.
This scenario simulates an expanding domain, where the model needs to continuously learn new areas without uncontrollably increasing its capacity. Our goal is to test whether the methods can form representations by progressively spreading inducing points and learning hyperparameters without prior tuning in two scenarios: 1) sequentially mapping multiple robot trajectories and 2) processing a single path in continuous batches.
\cref{fig:magneto_main} shows the final magnetic field estimates produced by VIPS in both experimental settings. Comparative results for CV and OIPS are provided in App.\,\ref{appendix:magneto}. The results show that VIPS was able to grow its memory as the robot moved, allowing for accurate modelling of the path and increasing its capacity dynamically. In contrast, we observe that CV increases its capacity uncontrollably while OIPS provides a less accurate estimate of the magnetic field due to choosing fewer inducing points.

\section{Discussion}
In this work, we propose a method to dynamically adjust the number of inducing variables in streaming GP regression. Our method achieves a performance close to full-batch approaches while minimising model size. It relies on a single hyperparameter to balance accuracy and complexity, and we demonstrate that our method performs consistently across diverse datasets. This reduces the need for extensive hyperparameter tuning and eliminates the requirement to pre-define model size, thereby addressing a significant bottleneck in traditional methods. 
While our current application focuses on GPs we hope this is extendable for the development of adaptive neural networks.

\section*{Acknowledgements}

GPB is supported by EPSRC through the Statistical Machine Learning (StatML) CDT programme, grant no. EP/S023151/1. GPB acknowledges the support of G-Research for her participation in the conference.

\section*{Impact Statement}

This paper presents work whose goal is to advance the field of 
Machine Learning. There are many potential societal consequences 
of our work, none which we feel must be specifically highlighted here.

\bibliography{bibliography} 
\bibliographystyle{icml2025}

\newpage
\appendix
\onecolumn
\section{Code}
The methods discussed in this work, along with the code to reproduce our results, are available online at \url{https://github.com/guiomarpescador/vips}. All experiments and figures were done in Python using libraries TensorFlow v.2.12 \citep{tensorflow} and GPflow v.2.9 \citep{GPflow2017}. Both packages are under Apache License 2.0. The implementation of the online lower bound (see eq.\,\eqref{eq:onlinecollapsed}) by \citet{Bui2017} is available on \href{https://github.com/thangbui/streaming_sparse_gp}{GitHub} under Apache license 2.0. For the ``greedy variance'' selection criterion and algorithm, we use the implementation provided in \citet{Burt2020} available under Apache license 2.0.

\section{Adaptive Inducing Points Selection Methods}\label{appendix:methods}

In sections \ref{sec:uci} and \ref{sec:magneto} we compare our method, VIPS, to two other adaptive approaches: Conditional Variance (CV) and  OIPS \citep{Galy-Fajou2021-tu}. The key distinction between these approaches lies in their selection strategy of inducing points. OIPS uses a unified selection process (Algorithm \ref{algo:uni}), where the choice of both location and number of inducing points is determined by a single metric. In particular, each point $\mathbf{x}_i$ in the training set $X_n$ is evaluated individually, and those that meet the threshold condition are added to the inducing set $\mathbf{Z}_n$. In contrast, both CV and VIPS use a decoupled selection process (Algorithm \ref{algo:decou}), which separates two decisions: where to place inducing points and how many to include. These methods iteratively select the best candidate for the next inducing point from a pool of potential locations, continuing until a stopping criterion is met, allowing for more control over the model size. VIPS and CV share the same greedy variance selection criterion but differ in their candidate pools and stopping criteria. Table \ref{table:methods} summarises the properties of these methods, including the fixed memory approach of \citet{Bui2017} for comparison. The next three subsections provide detailed descriptions of each method's specific implementation.

\begin{algorithm}[htpb]
\caption{Unified inducing set selection process done at each batch.}\label{algo:uni}
\begin{algorithmic}
\STATE {\bfseries Input:} New batch of data $\mathbf{X}_n$, current set inducing points $\mathbf{Z}_{old}$.
\STATE {\bfseries Output:} Updated set $\mathbf{Z}_{n}$
\STATE Initialise $\mathbf{Z}_n= \mathbf{Z}_{old}$
\FOR{$\mathbf{x}_{i} \in \mathbf{X}_{n}$} 
        \IF{addition\_criterion($\mathbf{x}_{i}$) < threshold}
\STATE $\mathbf{Z}_n= \mathbf{Z}_n\cup \{\mathbf{x}_{i}\}$
   \ENDIF
\ENDFOR
\end{algorithmic}
\end{algorithm}

\begin{algorithm}[htpb]
\caption{Decoupled inducing set selection process done at each batch.}\label{algo:decou}
\begin{algorithmic}
\STATE {\bfseries Input:} New batch of data $\mathbf{X}_n$, current set inducing points $\mathbf{Z}_{old}$.
 \STATE {\bfseries Output:} Updated set $\mathbf{Z}_{n}$
\STATE Set candidate pool $\mathbf{X}_{pool}$ based on method (see Table~\ref{table:methods}) and initialise $\mathbf{Z}_n$ accordingly.
\WHILE{stopping\_criterion($\mathbf{Z}_n$) $>$ threshold}
   \STATE $\mathbf{x}_{best} =$ selection\_criterion($\mathbf{X}_{pool}$, $\mathbf{Z}$)  \COMMENT{Greedy variance criterion}
   \STATE $\mathbf{Z}_n= \mathbf{Z}_n\cup \{\mathbf{x}_{best}\}$
\ENDWHILE
\end{algorithmic}
\end{algorithm}

\begin{table*}[htpb]
\caption{Properties of inducing points selection method for updating an online GP regression model.}
\begin{center}
\begin{small}
\begin{sc}
\begin{tabular}{lllll }
  Method & Model Size & Selection Pool &  Selection Criterion &  Stopping Criterion\\
\midrule
 \citet{Bui2017} &  Fixed & $\{\mathbf{Z}_o, \mathbf{X}_n\}$  & Gradient optimisation & M constant \\
 OIPS \citep{Galy-Fajou2021-tu}   &  Adaptive & $\{\mathbf{X}_n\}$ & $\max \mathbf{k}_\mathbf{u}(\mathbf{x}) \leq \rho$  &  $\max \mathbf{k}_\mathbf{u}(\mathbf{x}) \leq \bar\rho$  \\
Cond. Variance (CV) & Adaptive & $\{\mathbf{Z}_o, \mathbf{X}_n\}$ & Greedy variance &  $\operatorname{tr}\left(\mathbf{K_{ff}} - \mathbf{Q_{ff}} \right) \leq \eta$. \\
VIPS &  Adaptive & $\{\mathbf{X}_n\}$  & Greedy variance  &   $\widehat{\mathcal{U}} -\widehat{\mathcal{L}} \leq  \alpha$\\
\bottomrule
\end{tabular}
\end{sc}
\end{small}
\end{center}
\label{table:methods}
\end{table*}

\subsection{Vegas Inducing Point Selection (VIPS) Algorithm}\label{append:algo}

To select the location of our new inducing points we use the location selection strategy ``greedy variance'' proposed in \citet{Burt2020}. This strategy iteratively selects points from a set based on a preference criterion until a stopping condition is met. In particular, it chooses the location of the next inducing point to maximise the marginal variance in the conditional prior $p(f_{\neq\mathbf{u}}|\mathbf{u})$. This is equivalent to maximising $\operatorname{diag}[\mathbf{K_{ff}} - \mathbf{Q_{ff}} ]$. In continual learning, \citet{Chang2023} use the ``greedy variance'' criterion by defining $\{\mathbf{Z}_{o}, \mathbf{X}_{n}\}$ as the selection pool from which inducing point locations are selected and maintaining a fixed number of inducing points. Similarly, \citet{Maddox} extends the ``greedy variance'' criterion to heteroskedastic Gaussian likelihoods and also uses a fixed memory approach.  In our case, we tested the location strategy with our stopping criterion using both $\{\mathbf{Z}_{o}, \mathbf{X}_{n}\}$ and $\{\mathbf{X}_{n}\}$ as candidates pool for the locations of the inducing points. We did not find a substantial difference between the methods and hence opted for the simpler version where we keep the old inducing point locations fixed and choose the new set of inducing points from among the locations in $\{\mathbf{X}_{n}\}$.  

Algorithm \ref{alg:1} presents an inducing point selection method using our stopping criterion combined with the location selection strategy ``greedy variance''.
The method takes as input a value for the hyperparameter $\theta_{n}$. In practice, we will set $\theta_{n}=\theta_{o}$ to select the number of inducing points; the hyperparameter $\theta_{n}$ is inferred by optimising $\widehat{\mathcal{L}}(M_\mathbf{b})$ once the inducing point locations $\mathbf{Z}_{n}$ has been chosen. In Algorithm 1, $\widehat{\mathcal{U}}(M)$ is used to calculate the stopping criterion. However, in practice,  since for the continual learning settings we investigate $N_n$ is small enough,  we will use $\widehat{\mathcal{U}}(M_\mathbf{a} + N_n) = \mathcal{L}^*$. This value is calculated once at the beginning of the process. The algorithm's complexity depends on the number of inducing points $M_\mathbf{b}$ used to compute the lower bound $\widehat{\mathcal{L}}(M_\mathbf{b})$ at each iteration. The computational complexity for calculating $\widehat{\mathcal{L}}$ at each batch is $O(N_{n}M_\mathbf{b}^2 + M_\mathbf{b}^3)$, and the memory requirement is $O(M_\mathbf{b}^2)$, where $M_\mathbf{b}$ represents the total number of inducing points in the current batch.

\begin{algorithm}[htpb]
\caption{Vegas Inducing Point Selection (VIPS)}\label{alg:1}
\begin{algorithmic}
   \STATE {\bfseries Input:} $\mathbf{X}_{n} = \{\mathbf{x}_i\}_{i=1}^{N_{n}}$, $\mathbf{Z}_{o} =  \{\mathbf{z}_m\}_{m=1}^{M_{\mathbf{a}}}$, $\hat{\mu}$, $\hat\sigma$, $\theta_{n}$, kernel $k(\cdot, \cdot | \theta_{n})$, threshold parameter $\delta$. 
   \STATE {\bfseries Output:} Updated set \(\mathbf{Z}_{n} = \mathbf{Z}_{o} \cup \{\mathbf{x}_{m'}\}_{m'=1}^{M'}\), where \(|\mathbf{Z}_{n}| = M_\mathbf{b}\).
   \STATE Initialise $\mathbf{Z}_{n} =\mathbf{Z}_{o}$. 
   \WHILE{$\widehat{\mathcal{U}}(M)-\widehat{\mathcal{L}}(M_{\mathbf{b}}) \leq \delta|\widehat{\mathcal{U}}(M) - \mathcal{L}_{noise}(\hat{\mu}, \hat{\sigma})|$}
    \STATE Select $ \mathbf{x} = {\operatorname{argmax}}_{ \mathbf{x} \in \mathbf{X}_n}\, k(\mathbf{x}, \mathbf{x}) - \mathbf{k}_\mathbf{b}(\mathbf{x})^\top\mathbf{K}_\mathbf{bb}^{-1} \mathbf{k}_\mathbf{b}(\mathbf{x})$. 
   \STATE Add  $\mathbf{x}$ to the set of inducing points: $\mathbf{Z}_{n} = \mathbf{Z}_{n}\cup\{\mathbf{x}\}$.
   \ENDWHILE
\end{algorithmic}
\end{algorithm}

\subsection{Conditional Variance}
The implementation of the Conditional Variance method is presented in Algorithm \ref{alg:cv}. This method uses the ``greedy variance'' strategy that iteratively chooses the location of the next inducing point. As a stopping criterion, it uses the quantity $\operatorname{tr}\left(\mathbf{K_{ff}} - \mathbf{Q_{ff}} \right)$. In this algorithm, new inducing points are no longer added once $\operatorname{tr}\left(\mathbf{K_{ff}} - \mathbf{Q_{ff}} \right)$ falls below a chosen tolerance value $\eta$. Although, this approach was mentioned in \citet{Burt2020}, this stopping criterion has not yet been tested in the literature. The hyperparameter $\eta$ is determined by the user.

\begin{algorithm}[htpb]
\caption{Conditional Variance (CV)}\label{alg:cv}
\begin{algorithmic}
   \STATE {\bfseries Input:} $\mathbf{X}_{n} = \{\mathbf{x}_i\}_{i=1}^{N_{n}}$, $\mathbf{Z}_{o} =  \{\mathbf{z}_m\}_{m=1}^{M_{\mathbf{a}}}$, $\theta_{n}$, kernel, $k(\cdot, \cdot | \theta_{n})$, threshold $\eta$.
    \STATE {\bfseries Output:} Updated set of inducing points $\mathbf{Z}_{n} = \{\mathbf{z}_{m}\}_{m=1}^{M'} \cup \{\mathbf{x}_{m'}\}_{m'=1}^{M_b - M'}$, where \(|\mathbf{Z}_{n}| = M_\mathbf{b}\).
   \STATE Initialise location selection pool: $\mathbf{X}_{pool} =  \mathbf{Z}_{o}\cup\mathbf{X}_m$.
   \STATE Initialise $\mathbf{Z}_{n} = {\operatorname{argmax}}_{ \mathbf{x} \in \mathbf{X}_{pool}}\, k(\mathbf{x}, \mathbf{x})$. 
   \WHILE{$\operatorname{tr}\left(\mathbf{K_{ff}} - \mathbf{Q_{ff}} \right) \leq \eta$}
    \STATE Select $ \mathbf{x} = {\operatorname{argmax}}_{ \mathbf{x} \in \mathbf{X}_{pool}}\, k(\mathbf{x}, \mathbf{x}) - \mathbf{k}_\mathbf{b}(\mathbf{x})^\top\mathbf{K}_\mathbf{bb}^{-1} \mathbf{k}_\mathbf{b}(\mathbf{x})$. 
   \STATE Add $\mathbf{x}$ to the set of inducing points: $\mathbf{Z}_{n} = \mathbf{Z}_{n}\cup\{\mathbf{x}\}$.
   \ENDWHILE
\end{algorithmic}
\end{algorithm}

\subsection{Online Inducing Point Selection (OIPS)}

\citet{Galy-Fajou2021-tu} introduced the Online Inducing Points Selection (OIPS) algorithm, which iteratively adds points from $\mathbf{X}_{n}$ to the set of inducing points. The algorithm assesses the impact of each new point on the existing inducing set, based on a covariance threshold. A point $\mathbf{x}$ is added if the maximum value of $\mathbf{k}_\mathbf{u}(\mathbf{x})$ falls below a user-defined threshold $\rho$. While the original algorithm implicitly assumes unit kernel variance, Algorithm \ref{alg:oips} presents our implementation where we scale the threshold $\rho$ by the kernel variance $\sigma^2_f$ to $\bar\rho = \sigma^2_f\rho$. This modification ensures the threshold comparison remains valid for arbitrary kernel variances.

\begin{algorithm}[htpb]
\caption{Online Inducing Point Selection (OIPS)}\label{alg:oips}
\begin{algorithmic}
    \STATE {\bfseries Input:} $\mathbf{X}_{n} = \{\mathbf{x}_i\}_{i=1}^{N_{n}}$, $\mathbf{Z}_{o} =  \{\mathbf{z}_m\}_{m=1}^{M_{\mathbf{a}}}$, kernel function $k(\cdot, \cdot | \theta_{n})$, kernel hyperparameters $\theta_{n}$ (including variance $\sigma_f^2$), acceptance threshold $0<\rho<1$.
   \STATE {\bfseries Output:} Updated set \(\mathbf{Z}_{n} = \mathbf{Z}_{o} \cup \{\mathbf{x}_{m'}\}_{m'=1}^{M'}\), where \(|\mathbf{Z}_{n}| = M_\mathbf{b}\).
   \STATE Initialise $\mathbf{Z}_{n} =\mathbf{Z}_{o}$. 
   \STATE Initialise $\Bar{\rho} = \rho \cdot \sigma^2_f$.
    \FORALL{$\mathbf{x}_i \in \mathbf{X}_{n}$}
        \STATE $d = \max _j\left(k\left(\mathbf{x}_i, \mathbf{z}_j | \theta_{n}\right)\right), \, \forall\,\mathbf{z}_j \in \mathbf{Z}_{n}$.
        \IF{$d < \Bar{\rho}$}
            \STATE Add $\mathbf{x}_i$ to the set of inducing points: $\mathbf{Z}_{n} = \mathbf{Z}_{n}\cup\{\mathbf{x}_i\}$.
        \ENDIF
    \ENDFOR
\end{algorithmic}
\end{algorithm}

\section{Proof of Guarantee}\label{appendix:proof}

\begin{guarantee}
Let $M$ be a fixed integer and $M_\mathbf{b}$ be the number of selected inducing points such that $\widehat{\mathcal{U}}(M) - \widehat{\mathcal{L}}(M_\mathbf{b}) \leq \alpha$. Assuming that $\theta_n=\theta_o$, we have two equivalent bounds:
\begin{align}
\operatorname{KL}[q_{n}(f)\mid\mid p(f|\mathbf{y}_{o}, \mathbf{y}_{n}, \theta_{o})]&\leq \alpha + \Psi   \\
\mathrm{KL}[q_n(f)\mid\mid q_n^*(f)] &\leq \alpha
\end{align}
where $\Psi = \int q_n(f) \log \frac{q_n^*(f)}{p(f|\mathbf y_n, \mathbf y_o)} \mathrm d f$ and $q_n^*(f) = \mathcal{Z}^{-1} q_o(f) p(\mathbf{y}_n | f)$ represents the variational distribution associated with the optimal lower bound $\mathcal{L}^* = \widehat{\mathcal{L}}(N_n + M_{\mathbf{a}})$, with $\mathcal{Z}$ denoting the marginal likelihood that normalises $q_n^*(f)$.
\end{guarantee}

\begin{proof}
We cease the addition of points when $\widehat{\mathcal{U}}(M) - \widehat{\mathcal{L}}(M_\mathbf{b}) < \alpha $. Since $\widehat{\mathcal{U}}(M) \geq \mathcal{L}^*$, then $ - \widehat{\mathcal{L}}(M_\mathbf{b}) < \alpha - \widehat{\mathcal{U}}(M)  < \alpha - \mathcal{L}^*$. Eq.(4) can be bounded as: 
\begin{equation}
\begin{aligned}\label{eq:KLboundappendix}
\operatorname{KL}[q_{n}(f)\mid\mid p(f|\mathbf{y}_{o},\mathbf{y}_{n}, \theta_{n})] 
&= \log \frac{p(\mathbf{y}_{n}, \mathbf{y}_{o} | \theta_{n}) }{p(\mathbf{y}_{o} | \theta_{o}) }- \widehat{\mathcal{L}} + \Phi \\
&\leq \log \frac{p(\mathbf{y}_{n}, \mathbf{y}_{o} | \theta_{n}) }{p(\mathbf{y}_{o} | \theta_{o}) }+ \alpha - \mathcal{L}^* + \Phi \\
\end{aligned}
\end{equation}
where $\Phi = -\int q_{n}(f) \log  \frac{p(f|\mathbf{y}_{o}, \theta_{o})}{q_{o}(f)}\text{d}f$. Let $q_n^*(f) = \mathcal{Z}^{-1} q_o(f) p(\mathbf{y}_n | f)$ be the variational distribution associated with $\mathcal L^* = \widehat{\mathcal{L}}(N_n + M_{\bf a})$. Then,  by expanding the true posterior and multiplying by the variational distributions $q_n^*(f)$ on both sides of the fraction inside the log, we obtain: 
\begin{equation}
\begin{aligned}
\Phi&= \int q_{n}(f) \log \frac{q_{o}(f)}{p(f|\mathbf{y}_{o}, \theta_o)}\text{d}f\\
    &= \int q_{n}(f) \log \frac{q_{o}(f)p(\mathbf{y}_{o} |\theta_o)}{p(\mathbf{y}_{o} | f )p(f|\theta_o)}\text{d}f\\
    &= \int q_{n}(f) \log \frac{q_{o}(f)p(\mathbf{y}_{o} |\theta_o)}{p(\mathbf{y}_{o} | f )p(f|\theta_o)}\frac{q_n^*(f)}{q_n^*(f)}\text{d}f\\
    &= \int q_{n}(f)\log \frac{\cancel{q_{o}(f)}p(\mathbf{y}_{o} |\theta_o)}{p(\mathbf{y}_{o} | f )p(f|\theta_o)}\frac{q_n^*(f)}{\mathcal{Z}^{-1} \cancel{q_o(f)}p(\mathbf y_n|f)}\text{d}f\\
    &=  \int q_{n}(f)\log \frac{q_n^*(f)}{p(f|\mathbf{y}_{n}, \mathbf{y}_{o}, \theta_o)}\text{d}f + \log \mathcal{Z} -  \log  \frac{p(\mathbf{y}_{n}, \mathbf{y}_{o} | \theta_{o}) }{p(\mathbf{y}_{o} | \theta_{o}) } \\
\end{aligned}
\end{equation}
Using the above expansion for $\Phi$, eq.\,\eqref{eq:KLboundappendix} becomes, 
\begin{equation}
\begin{aligned}
&\operatorname{KL}[q_{n}(f)\mid\mid p(f|\mathbf{y}_{o},\mathbf{y}_{n}, \theta_{n})] \\
&\quad\leq \log  \frac{p(\mathbf{y}_{n}, \mathbf{y}_{o} | \theta_{n}) }{p(\mathbf{y}_{o} | \theta_{o}) } + \alpha - \mathcal{L}^* + \Phi \\
&\quad\leq \log  \frac{p(\mathbf{y}_{n}, \mathbf{y}_{o} | \theta_{n}) }{p(\mathbf{y}_{o} | \theta_{o}) } + \alpha - \mathcal{L}^* +\int q_{n}(f)\log \frac{q_n^*(f)}{p(f|\mathbf{y}_{n}, \mathbf{y}_{o}, \theta_{o})}\text{d}f + \log \mathcal{Z} - \log  \frac{p(\mathbf{y}_{n}, \mathbf{y}_{o} | \theta_{o}) }{p(\mathbf{y}_{o} | \theta_{o}) }.\\
\end{aligned}
\end{equation}
Assuming that $\theta_n = \theta_o$, the above can be simplified to
\begin{equation}
\operatorname{KL}[q_{n}(f)\mid\mid p(f|\mathbf{y}_{o},\mathbf{y}_{n}, \theta_{n})] \leq  \alpha +\int q_{n}(f)\log \frac{q_n^*(f)}{p(f|\mathbf{y}_{n}, \mathbf{y}_{o}, \theta_{n})}\text{d}f
\end{equation}
Again by multiplying by $q_{n}(f)$ both sides of the fraction inside the log, we obtain: 
\begin{equation}
\begin{aligned}
\operatorname{KL}[q_{n}(f)\mid\mid p(f|\mathbf{y}_{o},\mathbf{y}_{n}, \theta_{n})] &\leq 
 \int q_{n}(f)\log \frac{q_n^*(f)}{p(f|\mathbf{y}_{n}, \mathbf{y}_{o}, \theta_{n})}\frac{q_{n}(f)}{q_{n}(f)}\text{d}f + \alpha \\ 
 \operatorname{KL}[q_{n}(f)\mid\mid p(f|\mathbf{y}_{o},\mathbf{y}_{n}, \theta_{n})] &\leq  \operatorname{KL}[q_{n}(f)\mid\mid p(f|\mathbf{y}_{n},\mathbf{y}_{o}, \theta_{n})]  - \operatorname{KL}[q_{n}(f)\mid\mid q_{n}^*(f)] + \alpha \\ 
 \operatorname{KL}[q_{n}(f)\mid\mid q_{n}^*(f)] &\leq \alpha. \\
\end{aligned}
\end{equation}
\end{proof}

\section{Further Experimental Details and Results}\label{appendix:experiments}

 For all experiments and methods, we use the L-BFGS optimiser.

\subsection{Model size and data distribution}\label{appendix:datatypes}

For the synthetic dataset, we generate random noisy observations from the test function $f(x) = \sin(2x) + \cos(5x)$. We used a Squared Exponential kernel initialised with lengthscale $0.5$ and variance $1$. The noise variance was initialised to $0.5$.  For VIPS, we use $\delta = 0.05$. 

\paragraph{Dataset 1:} We use $N = 500$ observations uniformly distributed from $0$ to $10$. The data is ordered and divided into ten batches. 
\paragraph{Dataset 2:} We simulate a scenario where small batches of data are received but the data is distributed across the input space. We use $N = 150$ observations uniformly distributed from $0$ to $10$. The data is shuffled and divided into ten batches. 
\paragraph{Dataset 3:} We simulate a scenario where only outliers are encountered from time to time and the rest of the data is concentrated around a small part of the input space. We use two sets of data: the first set is sampled from a uniform distribution from $4$ to $6$, with $N=1000$ and the second set is sampled from a Cauchy distribution with a mean of $\mu=5$, with $N=300$. The data is divided into ten batches, where the first few batches only contain observations from the $4$ to $6$ range, and the Cauchy observations are observed in the latter batches.

\begin{figure}[htpb]
\vskip 0.2in
\centering
\includegraphics[width=\linewidth]{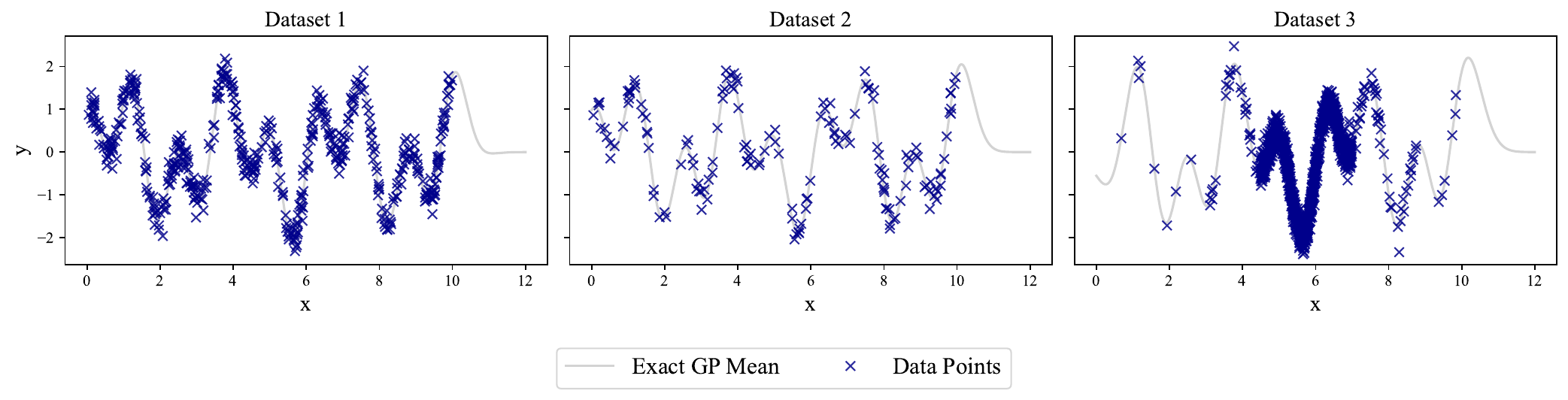}
    \caption{Plot of the three datasets considered in Section \ref{appendix:datatypes}.}
    \label{fig:syn}
\vskip -0.2in
\end{figure}

\subsection{The impact of model capacity in accuracy and training time}\label{appendix:accandtiming}
\subsubsection{Accuracy comparison}\label{appendix:synthetic}

For the synthetic dataset, we generate 1000 random noisy observations from the test function $f(x) = \sin(2x) + \cos(5x)$. We used a Squared Exponential kernel initialised with lengthscale $0.5$ and variance $1.0$. The noise variance was initialised to $0.5$. The performance was measured on a test grid of $500$ points. For VIPS, we use $\delta = 0.05$. 

\subsubsection{Training cost comparison}\label{appendix:timing}
This experiment was performed on an Nvidia RTX 6000's GPU on a high-performance computing cluster. We used a Squared Exponential kernel with hyperparameters initialised to 1. The noise variance was initialised to 0.1. The dataset was divided into 20 batches, and we recorded the time in training per batch. For VIPS, we use $\delta = 0.05$. 

\subsection{UCI datasets}\label{appendix:UCI}


These experiments were performed on an Nvidia RTX 6000's GPU on a high-performance computing cluster. We used a Squared Exponential kernel with hyperparameters initialised to 1 for all datasets. The noise variance was initialised to 0.1. 
We consider six UCI \citep{UCI} datasets of different characteristics: Concrete (1030, 8), Skillcraft (3338, 19), Kin8nm (8192, 8), Naval (11934, 14), Elevators (16599, 18), and Bike (17379, 17). The smaller datasets $(<12000)$ were divided into 20 batches, and the larger $(>12000)$ into 50 batches. The batches were created by sorting the data points by the first dimension.

Increasing the model size enhances performance until all relevant dataset information is captured; beyond this point, only computational costs increase. Therefore, when comparing inducing point methods, it is essential to consider the size-performance trade-off, rather than focusing solely on performance gains. This trade-off is typically controlled by a hyperparameter of the model. However, in continual learning, traditional cross-validation for tuning is not practical since it would require storing all past data. As a result, a good method needs to use pre-set hyperparameters and still perform well on different datasets.  We compare our method, VIPS, with two other adaptive approaches: Conditional Variance (CV) and OIPS. Our goal is to determine if a single hyperparameter for each method can perform consistently well across different datasets. To do this, we evaluate various hyperparameter settings for each method, with ranges $\delta \in [0.005, 0.2]$ for VIPS,  $\delta \in [0.005, 1.0]$ for CV, $\rho \in [0.80, 0.999]$ for OIPS. 

For each method, we identified hyperparameter values that achieved a particular accuracy threshold (either RMSE or NLPD) within a percentage of 5\% or 10\% of the full-batch GP relative to a noise model. For example, for 5\% RMSE, the accuracy threshold is calculated as: $\mathrm{RMSE}_\mathrm{exact} + 0.05\left|\mathrm{RMSE}_{\mathrm{noise}} - \mathrm{RMSE}_{\mathrm{exact}}  \right|$. Among the hyperparameter values meeting these criteria, we select the one that results in the smallest model size. For CV and VIPS, this corresponds to the largest hyperparameter value, while for OIPS, it corresponds to the smallest hyperparameter value (see Table \ref{table:methods} for a summary of the methods).

With the optimal hyperparameter selected, all methods achieve the desired performance threshold. Therefore, we can compare the methods based on their model size, where smaller sizes are preferred. As explained in the main part of the paper, we did not find a hyperparameter that satisfied the accuracy thresholds for OIPS in the Naval dataset. For this method, we selected the hyperparameter configuration that satisfied our conditions for five out of six datasets, treating Naval as an outlier. This decision was made considering that OIPS already selected larger models with this less stringent hyperparameter (for example 80\% of the data points for the Kin8nm dataset) and extending the hyperparameter range would have led to even larger dataset sizes.

Since such dataset-specific limitations are common in practical applications, and to ensure a fair comparison across methods, we evaluated configurations that allowed for either one or two datasets to be outliers for each method.  To report performance, we calculated the difference with the full-batch GP metric relative to a noise model. For each dataset and batch, we computed the relative RMSE percentage as $$\mathrm{RMSE}\,\% =\left( \frac{\mathrm{RMSE}_{\mathrm{method}} - \mathrm{RMSE}_{\mathrm{exact}}}{\left|\mathrm{RMSE}_{\mathrm{noise}}- \mathrm{RMSE}_{\mathrm{exact}}  \right|} \right) \times 100
$$ and similarly for the NLPD metric. The test set for each batch consists only of data from the current and previous batches.

\begin{figure}[htbp]
\vskip 0.2in
    \centering
    \begin{subfigure}[t]{0.49\linewidth}
        \centering
        \includegraphics[width=\linewidth]{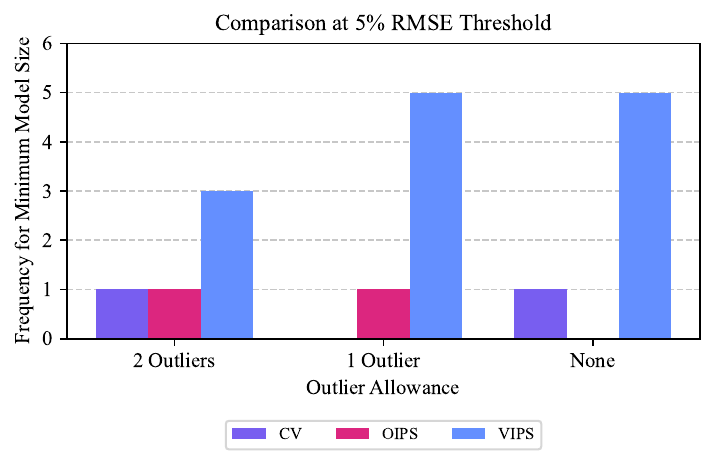}
        \caption{}
        \label{fig:sub1}
    \end{subfigure}
    \begin{subfigure}[t]{0.49\linewidth}
        \centering
        \includegraphics[width=\linewidth]{figures/method_comparison_10_RMSE.pdf}
        \caption{}
        \label{fig:sub2}
    \end{subfigure}

    \vspace{0.5cm} 

    \begin{subfigure}[t]{0.49\linewidth}
        \centering
        \includegraphics[width=\linewidth]{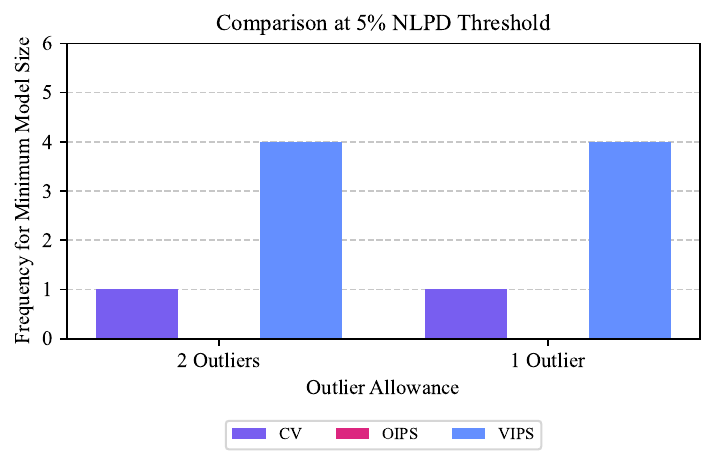}
        \caption{}
        \label{fig:sub3}
    \end{subfigure}
    \begin{subfigure}[t]{0.49\linewidth}
        \centering
        \includegraphics[width=\linewidth]{figures/method_comparison_10_NLPD.pdf}
        \caption{}
        \label{fig:sub4}
    \end{subfigure}
    \caption{Number of datasets where each method achieves minimal model size at different outlier allowances. A ``win'' is assigned to the dataset with the smallest model which satisfies the accuracy thresholds. Counts represent absolute wins. Higher counts indicate better method robustness across datasets.
}
\vskip -0.2in
    \label{fig:two_by_two}
\end{figure}

Figure \ref{fig:two_by_two} shows a summary of the results, presenting the absolute counts for smallest model size. Tables \ref{table:rmse5}, \ref{table:rmse10}, \ref{table:nlpd5} and \ref{table:nlpd10} provide details for each threshold and outlier choice. The results reveal consistent patterns across different accuracy thresholds and metrics. VIPS achieves the smallest model size while maintaining performance within thresholds, particulary when all datasets are considered. 

Allowing outlier datasets reduces model sizes across all methods. This comes with performance degradation on the particular outlier. We observe that this degradation is often higher for CV and OIPS. The Naval dataset proved particularly challenging for these methods. This is probably due to its almost noiseless nature. While VIPS came close (missing the threshold by 0.48\%), none of the methods achieved the 5\% target NLPD threshold within the tested hyperparameter range. Given the dataset's near-zero noise characteristics, we chose not to expand the hyperparameter range to avoid excessive model sizes.

The results for OIPS and CV suggest that their hyperparameters may need to be tuned for each dataset. As mentioned, this is not possible in continual learning where data is not available in advance. In contrast, across all configurations, VIPS demonstrates consistent behaviour, minimising model size in the majority of cases and hence reducing the need for dataset-specific tuning. Based on our experimental results, we recommend using $\delta = 0.035$ for VIPS, which represents the average of successful parameter values when all datasets were considered. This value should provide near-exact performance across different scenarios while minimising model size.

\newpage
\begin{table}[htbp]
\caption{Mean (std) of inducing points and RMSE \% difference with full-batch GP for the final batch at different outlier allowances at the selected operating point to achieve a 5\% RMSE threshold.  \xmark\,\textcolor{gray}{Gray} highlights cases that exceed the threshold, and \textbf{bold} text identifies the smallest model size. `Max.' denotes reaching the model's maximum capacity limit. For none, one, and two outliers, the selected hyperparameters were: VIPS $\delta = 0.02, 0.065, 0.14$, CV $\eta = 0.005, 0.06, 0.5$ and  OIPS $\rho = -, 0.955, 0.935$.}
\vskip 0.15in
\begin{center}
\begin{small}
\begin{sc}
\begin{tabular}{|c|c|cc|cc|cc|}
\hline
Dataset & Outliers & \multicolumn{2}{c|}{VIPS} & \multicolumn{2}{c|}{CV} & \multicolumn{2}{c|}{OIPS} \\
& & M & RMSE \% & M & RMSE \% & M & RMSE \% \\
\hline
\multirow{3}{*}{Concrete} 
& None & \textbf{434(79)} & 3.67\% & 492(54) & 3.28\% & - & -  \\
& One & \textbf{317(101)} & 4.24\% & 321(71) & 4.65\% & 383(90) & 4.89\% \\
& Two & \xmark \color{gray} 163(95) &\xmark  \color{gray} 10.43\% & \xmark\,\textcolor{gray}{70(5)} &\xmark\,\textcolor{gray}{ 13.34\%} & \xmark\,\textcolor{gray}{257(106)} & \xmark\,\textcolor{gray}{8.84\%} \\
\hline
\multirow{3}{*}{Skillcraft}
& None & \textbf{177(13)} & 0.14\% & 739(34) & 0.02\% & - & - \\
& One & \textbf{135(1)} & 0.21\% & 312(15) & 0.04\% & 236(38) & 0.07\% \\
& Two & 133(0) & 0.13\% & \textbf{130(56)} & 1.45\% & 185(21) & 0.02\% \\
\hline
\multirow{3}{*}{Kin8nm}
& None & \textbf{4281(69)} & 0.15\% & 6316(9) & 0.11\% & - & - \\
& One & \textbf{2513(54)} & 1.35\% & 5685(26) & 0.13\% & 6523(9) & 0.12\% \\
& Two & \textbf{1356(42)} & 4.97\% & 5334(27) & 0.13\% & 6499(10) & 0.12\% \\
\hline
\multirow{3}{*}{Naval}
& None & 373(19) & 0.40\% & \textbf{49(1)} & 2.39\% & - & - \\
& One & \textbf{86(3)} & 1.37\% & \xmark\,\textcolor{gray}{23(7)} & \xmark\,\textcolor{gray}{79.08\%} &\xmark\,\textcolor{gray}{ 20(1)}& \xmark\,\textcolor{gray}{407.22\%} \\
& Two & \textbf{43(2)} & 2.67\% & \xmark\,\textcolor{gray}{21(5)} & \xmark\,\textcolor{gray}{336.08\%} & \xmark\,\textcolor{gray}{20(1)}  & \xmark\,\textcolor{gray}{407.22\%} \\
\hline
\multirow{3}{*}{Elevators}
& None & \textbf{438(17)} & 1.19\% & 3133(140) & 0.31\% & - & - \\
& One & \textbf{312(5)} & 1.33\% & 1194(42) & 0.85\% & 386(47) & 1.19\% \\
& Two & \textbf{276(4)} & 1.55\% & 536(22) & 1.43\% & 305(13) & 1.49\% \\
\hline
\multirow{3}{*}{Bike}
& None & \textbf{2296(138)} & 4.61\% & Max. 7000 & 0.58\% & - & - \\
& One & \xmark\,\textcolor{gray}{853(19)} &\xmark\,\textcolor{gray}{ 8.73\%} & 5563(36) & 1.02\% & \textbf{3916(122)} & 2.19\% \\
& Two & \xmark\,\textcolor{gray}{498(10)} & \xmark\,\textcolor{gray}{11.24\%} & 3940(41) & 1.80\% &  \textbf{2438(130)} & 4.23\% \\
\hline
\end{tabular}
\end{sc}
\end{small}
\end{center}
\label{table:rmse5}
\end{table}

\begin{table}[H]
\caption{Mean (std) of inducing points and RMSE \% difference with full-batch GP for the final batch at different outlier allowances at the selected operating point to achieve a 10\% RMSE threshold.  \xmark\,\textcolor{gray}{Gray} highlights cases that exceed the threshold, and \textbf{bold} text identifies the smallest model size. `Max.' denotes reaching the model's maximum capacity limit. For none, one, and two outliers, the selected hyperparameters were: VIPS $\delta = 0.095, 0.12, 0.2$, CV $\eta = 0.005, 0.13, 0.6$ and  OIPS $\rho = -, 0.93, 0.845$.}
\vskip 0.15in
\begin{center}
\begin{small}
\begin{sc}
\begin{tabular}{|c|c|cc|cc|cc|}
\hline
Dataset & Outliers & \multicolumn{2}{c|}{VIPS} & \multicolumn{2}{c|}{CV} & \multicolumn{2}{c|}{OIPS} \\
& & M & RMSE \% & M & RMSE \% & M & RMSE \% \\
\hline
\multirow{3}{*}{Concrete} 
& None & \textbf{234(116)} & 7.66\% & 492(54) & 3.28\% & - & - \\
& One & \textbf{178(95)} & 9.81\% & 184(74) & 9.96\% & 240(102) & 9.03\% \\
& Two & \xmark\,\textcolor{gray}{111(37)} & \xmark\,\textcolor{gray}{12.93\%} & \xmark\,\textcolor{gray}{3(2)} & \xmark\,\textcolor{gray}{126.81\%} & \xmark\,\textcolor{gray}{108(66)} & \xmark\,\textcolor{gray}{20.79\%} \\
\hline
\multirow{3}{*}{Skillcraft}
& None & \textbf{134(0)} & 0.12\% & 739(34) & -0.02\% & - & - \\
& One & \textbf{133(0)} & 0.13\% & 226(22) & 0.20\% & 174(16) & 0.04\% \\
& Two & 133(0)& 0.12\% & \textbf{17(0)} & 7.93\% & 138(1) & 0.06\% \\
\hline
\multirow{3}{*}{Kin8nm}
& None & \textbf{1904(51)} & 2.55\% & 6316(9) & 0.11\% & - & - \\
& One & \textbf{1563(39)} & 3.77\% & 5334(27) & 0.13\% & 6458(8) & 0.11\% \\
& Two & \textbf{949(25)} & 8.17\% & 3408(32) & 0.19\% & 5053(93) & 0.14\% \\
\hline
\multirow{3}{*}{Naval}
& None & 57(1) & 2.25\% & \textbf{49(1)} & 2.39\% & - & - \\
& One & \textbf{48(3)} & 2.57\% & \xmark\,\textcolor{gray}{23(8)} & \xmark\,\textcolor{gray}{323.90\%} & \xmark\,\textcolor{gray}{16(1)} & \xmark\,\textcolor{gray}{315.28\%} \\
& Two & \textbf{38(1)} & 5.07\% & \xmark\,\textcolor{gray}{19(3)} & \xmark\,\textcolor{gray}{553.11\%} & \xmark\,\textcolor{gray}{13(1)} & \xmark\,\textcolor{gray}{407.6\%} \\
\hline
\multirow{3}{*}{Elevators}
& None & \textbf{291(4)} & 1.41\% & 3133(140) & 0.31\% & - & - \\
& One & \textbf{281(4)} & 1.51\% & 836(27) & 1.11\% & 298(8) & 1.49\% \\
& Two & \textbf{267(2)} & 1.57\% & 287(18) & 2.52\% & 269(1) & 1.59\% \\
\hline
\multirow{3}{*}{Bike}
& None & \textbf{650(17)} & 9.98\% & Max.\,7000 & 0.58\% & - & - \\
& One & \xmark\,\textcolor{gray}{549(14)} & \xmark\,\textcolor{gray}{10.82\%} & 4895(28) & 1.29\% & \textbf{1964(137)} & 5.18\% \\
& Two & \xmark\,\textcolor{gray}{400(7)} & \xmark\,\textcolor{gray}{12.50\%} & 2582(66) & 4.22\% & \textbf{594(11)} & 10.00\% \\
\hline
\end{tabular}
\end{sc}
\end{small}
\end{center}
\label{table:rmse10}
\end{table}

\begin{table}[htbp]
\caption{Mean (std) of inducing points and NLPD \% difference with full-batch GP for the final batch at different outlier allowances at the selected operating point to achieve a 5\% NLPD threshold.  \xmark\,\textcolor{gray}{Gray} highlights cases that exceed the threshold, and \textbf{bold} text identifies the smallest model size. `Max.' denotes reaching the model's maximum capacity limit. For one, and two outliers, the selected hyperparameters were: VIPS $\delta = 0.005, 0.015$, CV $\eta = 0.025, 0.100$ and  OIPS $\rho = 0.975, 0.960$.
}
\vskip 0.15in
\begin{center}
\begin{small}
\begin{sc}
\begin{tabular}{|c|c|cc|cc|cc|}
\hline
Dataset & Outliers & \multicolumn{2}{c|}{VIPS} & \multicolumn{2}{c|}{CV} & \multicolumn{2}{c|}{OIPS} \\
& & M & NLPD \% & M & NLPD \% & M & NLPD \% \\
\hline
\multirow{2}{*}{Concrete} 
& One & 514(74) & 2.27\% &\textbf{ 417(63)} & 4.28\% & 490(73) & 4.72\% \\
& Two & \textbf{451(77)} & 4.43\% & \xmark\,\textcolor{gray}{227(84)} & \xmark\,\textcolor{gray}{15.84\%} & \xmark\,\textcolor{gray}{410(88)} & \xmark\,\textcolor{gray}{8.34\%} \\
\hline
\multirow{2}{*}{Skillcraft}
& One & \textbf{291(34)} & 0.60\% & 439(29) & 0.60\% & 488(119) & 0.53\% \\
& Two & \textbf{195(18)} & 0.78\% & 260(16) & 0.81\% & 332(82) & 0.51\% \\
\hline
\multirow{2}{*}{Kin8nm}
& One & \textbf{5640(73)} & 0.85\% & 6194(12) & 0.88\% & 6547(8) & 0.88\% \\
& Two & \textbf{5065(46)} & 0.94\% & 5334(27) & 0.90\% & 6539(8) & 0.88\% \\
\hline
\multirow{2}{*}{Naval}
& One & \xmark\,\textcolor{gray}{1389(18)} & \xmark\,\textcolor{gray}{5.48\%} & \xmark\,\textcolor{gray}{27(5)} & \xmark\,\textcolor{gray}{4654.12\%} & \xmark\,\textcolor{gray}{26(1)} & \xmark\,\textcolor{gray}{115.29\%} \\
& Two & \xmark\,\textcolor{gray}{509(15)} & \xmark\,\textcolor{gray}{9.99\%} & \xmark\,\textcolor{gray}{28(4)} & \xmark\,\textcolor{gray}{8146.44\%} & \xmark\,\textcolor{gray}{20(1)} & \xmark\,\textcolor{gray}{133.87\%} \\
\hline
\multirow{2}{*}{Elevators}
& One & \textbf{740(24)} & 1.74\% & 1814(88) & 1.23\% & 1869(257) & 1.34\% \\
& Two & \textbf{487(24)} & 2.02\% & 916(21) & 1.91\% & 642(133) & 2.01\% \\
\hline
\multirow{2}{*}{Bike}
& One & \textbf{5064(334)} & 3.99\% & Max.\,7000 & 2.13\% & 6794(79) & 2.55\% \\
& Two &  \xmark\,\textcolor{gray}{2794(148)} & \xmark\,\textcolor{gray}{8.18\%} & 4895(28) & 3.54\% & 5131(64) & 4.05\% \\
\hline
\end{tabular}
\end{sc}
\end{small}
\end{center}
\label{table:nlpd5}
\end{table}

\begin{table}[htbp]
\caption{Mean (std) of inducing points and NLPD \% difference with full-batch GP for the final batch at different outlier allowances at the selected operating point to achieve a 10\% NLPD threshold.  \xmark\,\textcolor{gray}{Gray} highlights cases that exceed the threshold, and \textbf{bold} text identifies the smallest model size. `Max.' denotes reaching the model's maximum capacity limit. For one, and two outliers, the selected hyperparameters were: VIPS $\delta = 0.015, 0.02, 0.05$, CV $\eta = 0.001, 0.07, 0.5$ and  OIPS $\rho = -, 0.95, 0.935$.}
\vskip 0.15in
\begin{center}
\begin{small}
\begin{sc}
\begin{tabular}{|c|c|cc|cc|cc|}
\hline
Dataset & Outliers & \multicolumn{2}{c|}{VIPS} & \multicolumn{2}{c|}{CV} & \multicolumn{2}{c|}{OIPS} \\
& & M & NLPD \% & M & NLPD \% & M & NLPD \% \\
\hline
\multirow{3}{*}{Concrete} 
& None & \textbf{451(77)} & 4.43\% & 492(54) & 2.32\% & - & - \\
& One & 434(79) & 5.68\% & \textbf{303(78)} & 9.64\% & 383(90) & 9.56\% \\
& Two & \textbf{349(86)} & 9.04\% & \xmark\,\textcolor{gray}{3(2)} & \xmark\,\textcolor{gray}{118.71\%} & \xmark\,\textcolor{gray}{247(101)} & \xmark\,\textcolor{gray}{15.73\%} \\
\hline
\multirow{3}{*}{Skillcraft}
& None & \textbf{195(18)} & 0.78\% & 1091(107) & -0.34\% & - & - \\
& One & \textbf{177(13) }& 0.78\% & 295(13) & 0.71\% & 236(38) & 0.57\% \\
& Two & \textbf{141(4)} & 1.10\% & \xmark\,\textcolor{gray}{54(45)} & \xmark\,\textcolor{gray}{8.03\%} & 185(21) & 0.68\% \\
\hline
\multirow{3}{*}{Kin8nm}
& None & \textbf{5065(46)} & 0.94\% & 6469(7) & 0.87\% & - & - \\
& One & \textbf{4281(69)} & 0.34\% & 5685(26) & 0.91\% & 6523(9) & 0.88\% \\
& Two & \textbf{2953(71)} & 1.89\% & 5334(27) & 0.90\% & 6499(10) & 0.88\% \\
\hline
\multirow{3}{*}{Naval}
& None & \textbf{509(15)} & 9.99\% & 903(8) & 6.17\% & - & - \\
& One & \xmark\,\textcolor{gray}{373(19)} & \xmark\,\textcolor{gray}{11.92\%} & \xmark\,\textcolor{gray}{27(6)} & \xmark\,\textcolor{gray}{2405.87\%} & \xmark\,\textcolor{gray}{20(1)} & \xmark\,\textcolor{gray}{133.87\%} \\
& Two & \xmark\,\textcolor{gray}{126(6)} & \xmark\,\textcolor{gray}{22.95\%} & \xmark\,\textcolor{gray}{22(7)} & \xmark\,\textcolor{gray}{6012.82\%} & \xmark\,\textcolor{gray}{20(1)} & \xmark\,\textcolor{gray}{133.87\%} \\
\hline
\multirow{3}{*}{Elevators}
& None & \textbf{487(24)} & 2.02\% & 3133(140) & 0.82\% & - & - \\
& One & 438(17) & 2.11\% & 1099(50) & 1.73\% & 386(47) & 2.68\% \\
& Two & 332(8) & 2.54\% & 352(29) & 3.76\% & \textbf{305(13)} & 4.84\% \\
\hline
\multirow{3}{*}{Bike}
& None & \textbf{2794(148)} & 8.18\% & Max.\,7000 & 2.04\% & - & - \\
& One & \textbf{2296(138) }& 9.92\% & 5563(36) & 2.95\% & 3916(122) & 5.50\% \\
& Two &  \xmark\,\color{gray}1037(23) & \xmark\,\textcolor{gray}{17.16\%} & 2582(66) & 9.61\% & \textbf{2438(130)} & 9.66\% \\
\hline
\end{tabular}
\end{sc}
\end{small}
\end{center}
\label{table:nlpd10}
\end{table}

\newpage
\subsection{Magnetic anomalies}\label{appendix:magneto}

The data used in this experiment is obtained from \citet{Solin2018} and is available on \hyperlink{https://github.com/AaltoML/magnetic-data}{GitHub}. The objective of this task is to detect local anomalies in the Earth's magnetic field online, caused by the presence of bedrock and magnetic materials in indoor building structures. For this purpose, a small robot with a 3-axis magnetometer moves around an indoor space of approximately 6 meters by 6 meters and measures the magnetic field strength. Out of the 9 available trajectories, we use trajectories 1, 2, 3, 4, and 5 (with $n=8875,9105,9404,7332,8313$, respectively) for the experiments. Specifically, we use trajectories 1, 2, 4 and 5 for the first experiment and trajectory 3 for the second. 

We use the experimental setup proposed in \citet{Chang2023}.  The proposed model applies a GP prior to magnetic field strength, given by $\mathcal{G} \mathcal{P}\left(0, \sigma_0^2+\kappa_{\sigma^2, \ell}^{\mathrm{Mat}}\left(\mathbf{x}, \mathbf{x}^{\prime}\right)\right)$ (in $\mu \mathrm{T}$ ), where the kernel consist of a constant kernel and a Matérn-$\nu / 2$ kernel. The model assumes the spatial domain is affected by Gaussian noise with a variance $\sigma_{\mathrm{n}}^2$. The initial variance for the constant kernel is set to $500$, and the Gaussian likelihood is initialised with a noise variance of $0.1$. 

Our aim for this experiment is to test the optimal hyperparameters identified in the previous section for the 10\% RMSE threshold for each adaptive method in a real-world setting. The setting simulates an ever-expanding domain, where the robot is not confined to a predefined area. In this context, the model continuously learns new parts of the space. Therefore, a method that works will need to sufficiently expand the model’s size to accommodate new data without letting it grow uncontrollably.

\begin{figure}[htpb]
\begin{center}
    \begin{subfigure}{0.24\columnwidth}
        \centering
        \includegraphics[width=\linewidth]{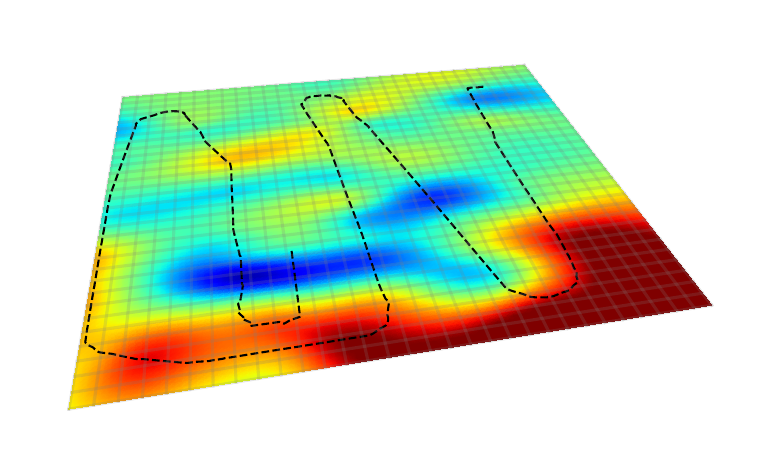} 
        \caption*{After path \#1}
    \end{subfigure}%
    \begin{subfigure}{0.24\columnwidth}
        \centering
        \includegraphics[width=\linewidth]{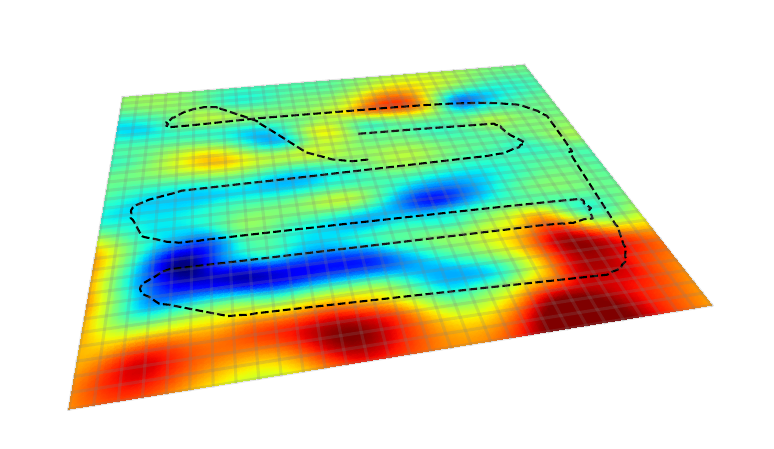} 
        \caption*{After path \#2}
    \end{subfigure}
    \begin{subfigure}{0.24\columnwidth}
        \centering
        \includegraphics[width=\linewidth]{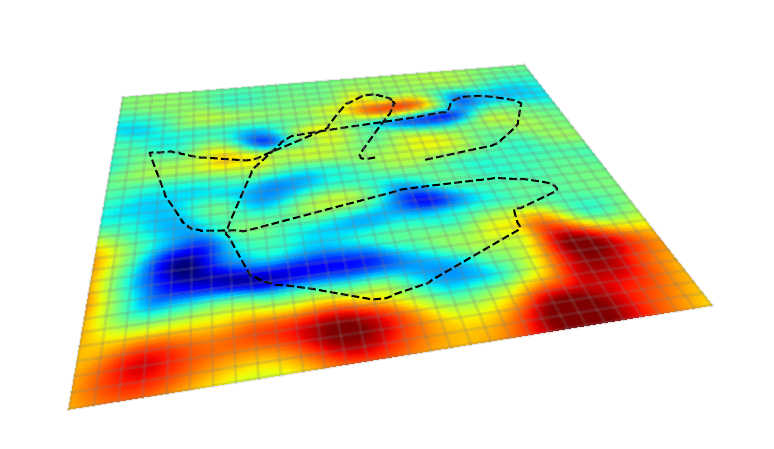} 
        \caption*{After path \#4}
    \end{subfigure}
    \begin{subfigure}{0.24\columnwidth}
        \centering
        \includegraphics[width=\linewidth]{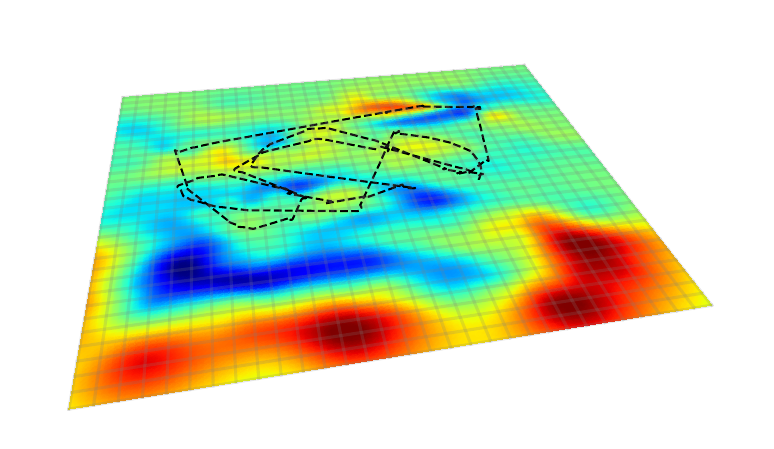} 
        \caption*{After path \#5}
    \end{subfigure}%
\caption{VIPS (Ours). A small robot with wheels is used to perform sequential estimation of magnetic field anomalies. We show the estimate of the magnitude field learned sequentially after travelling the path shown in a dotted line. The degree of transparency represents the marginal variance.
}
\label{fig:magneto_paths}
\end{center}
\end{figure}

\begin{figure}[htpb]
\begin{center}
    \begin{subfigure}{0.24\columnwidth}
        \centering
        \includegraphics[width=\linewidth]{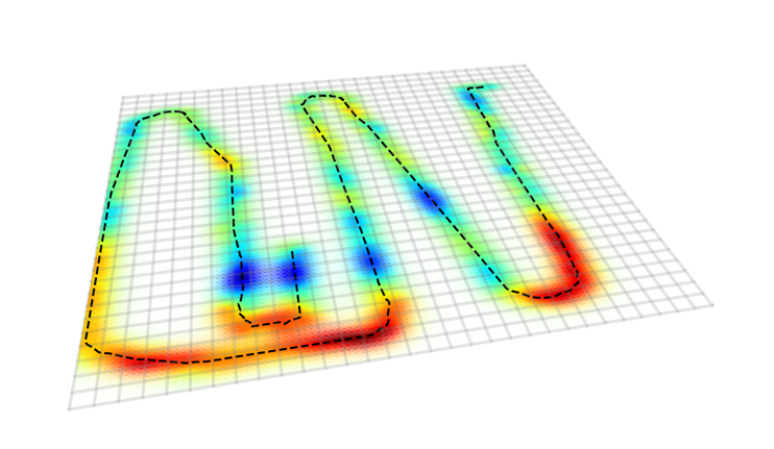} 
        \caption*{After path \#1}
    \end{subfigure}%
    \begin{subfigure}{0.24\columnwidth}
        \centering
        \includegraphics[width=\linewidth]{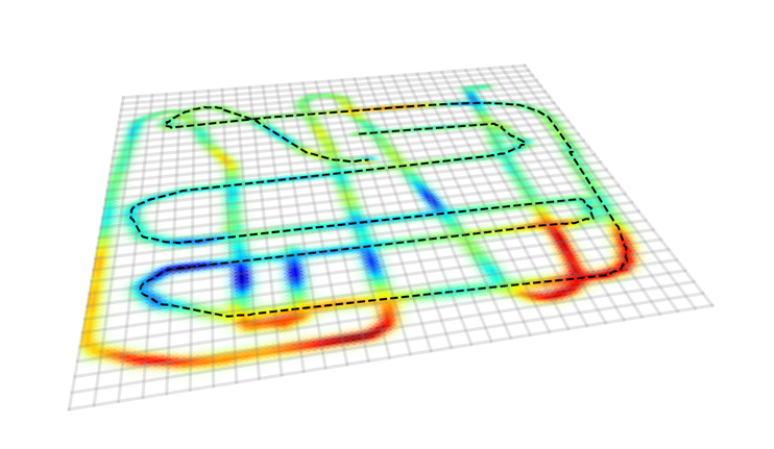} 
        \caption*{After path \#2}
    \end{subfigure}
    \begin{subfigure}{0.24\columnwidth}
        \centering
        \includegraphics[width=\linewidth]{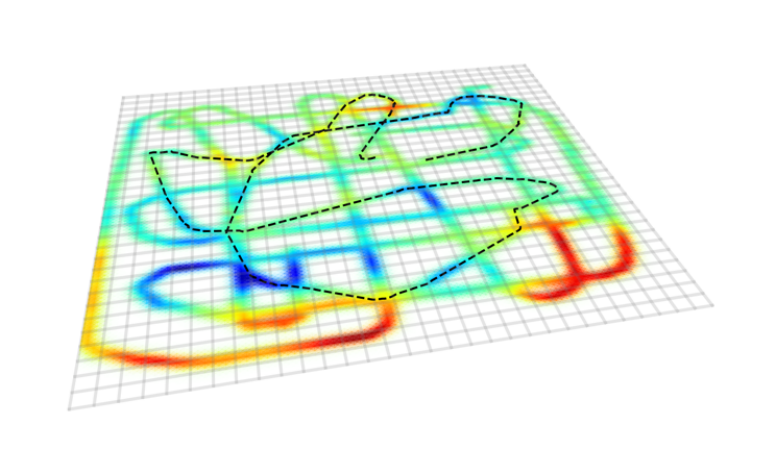} 
        \caption*{After path \#4}
    \end{subfigure}
    \begin{subfigure}{0.24\columnwidth}
        \centering
        \includegraphics[width=\linewidth]{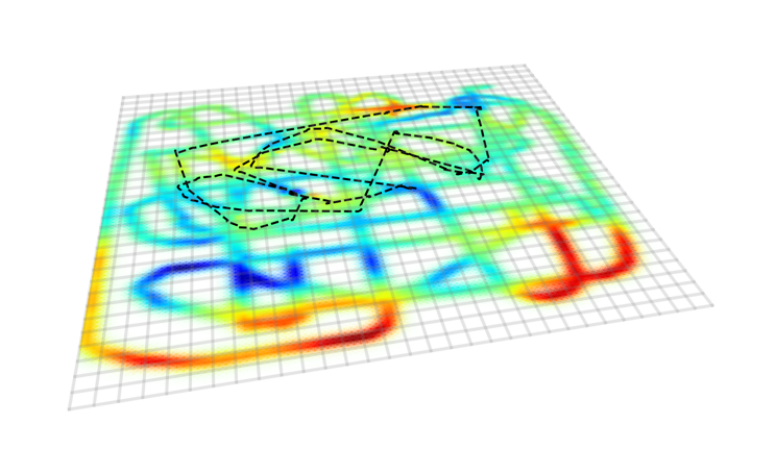} 
        \caption*{After path \#5}
    \end{subfigure}%
\caption{Conditional Variance. A small robot with wheels is used to perform sequential estimation of magnetic field anomalies. We show the estimate of the magnitude field learned sequentially after travelling the path shown in a dotted line. The degree of transparency represents the marginal variance.
}
\label{fig:cv_magneto_paths}
\end{center}
\end{figure}

\begin{figure}[htpb]
\begin{center}
    \begin{subfigure}{0.24\columnwidth}
        \centering
        \includegraphics[width=\linewidth]{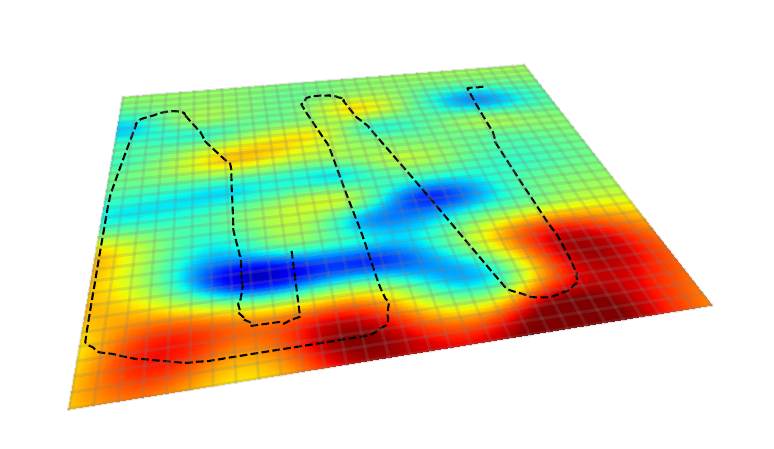} 
        \caption*{After path \#1}
    \end{subfigure}%
    \begin{subfigure}{0.24\columnwidth}
        \centering
        \includegraphics[width=\linewidth]{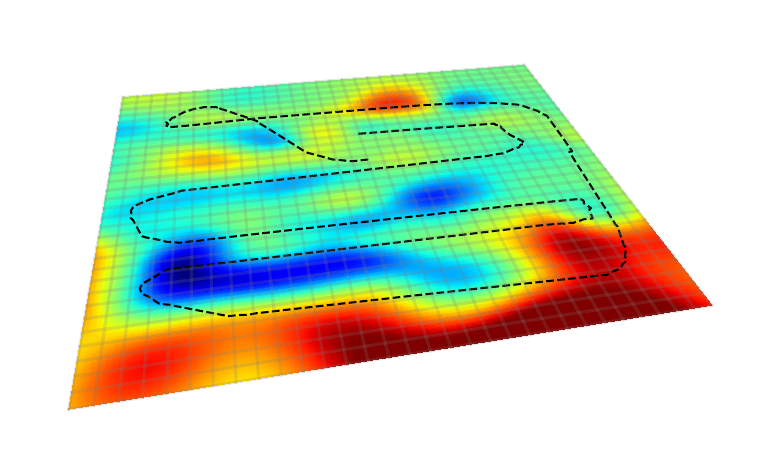} 
        \caption*{After path \#2}
    \end{subfigure}
    \begin{subfigure}{0.24\columnwidth}
        \centering
        \includegraphics[width=\linewidth]{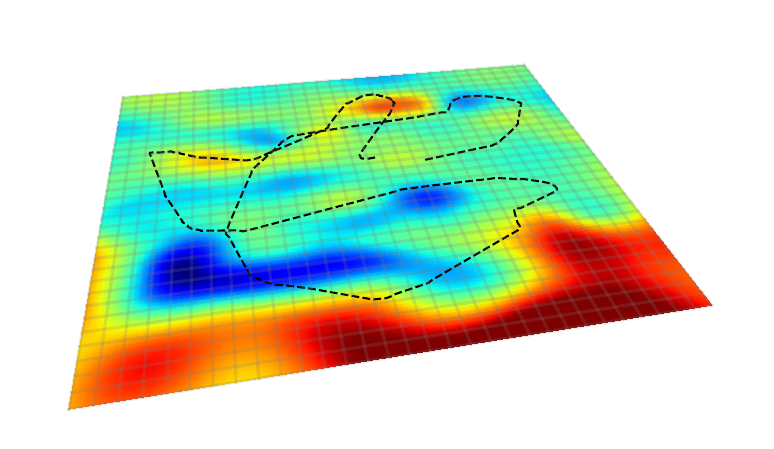} 
        \caption*{After path \#4}
    \end{subfigure}
    \begin{subfigure}{0.24\columnwidth}
        \centering
        \includegraphics[width=\linewidth]{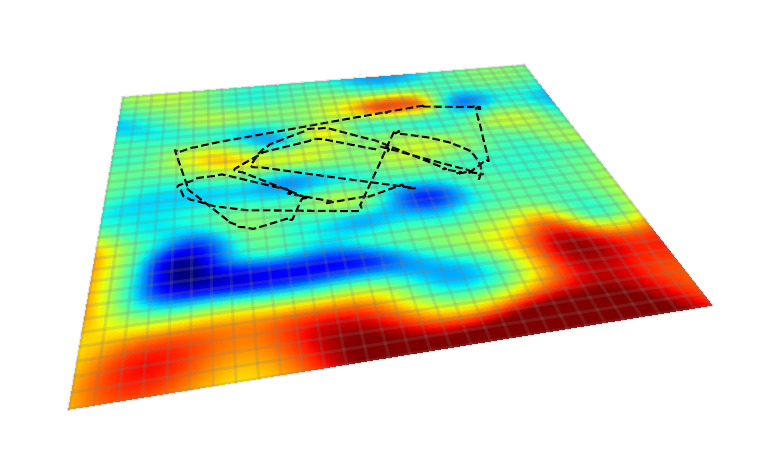} 
        \caption*{After path \#5}
    \end{subfigure}%
\caption{OIPS. A small robot with wheels is used to perform sequential estimation of magnetic field anomalies. We show the estimate of the magnitude field learned sequentially after travelling the path shown in a dotted line. The degree of transparency represents the marginal variance.
}
\label{fig:oips_magneto_paths}
\end{center}
\end{figure}

In the first experiment, we aim to sequentially learn the paths taken by the robot using trajectories 1, 2, 4 and 5, i.e. an entire path will correspond to one batch. We investigate whether the method can adapt to changes in the environment and adjust the number of inducing points accordingly. During this process, we concurrently learn the hyperparameters $\sigma_0^2$, $\sigma^2$, $\ell$, and $\sigma_{\mathrm{n}}^2$. As a test set, we use trajectory 3. Figures \ref{fig:magneto_paths}, \ref{fig:cv_magneto_paths}, \ref{fig:oips_magneto_paths} show the temporally updating field estimate over batches alongside the corresponding path travelled in each batch. 

\begin{figure}[htpb]
\begin{center}
    \begin{subfigure}{0.19\linewidth}
        \centering
        \includegraphics[width=\linewidth]{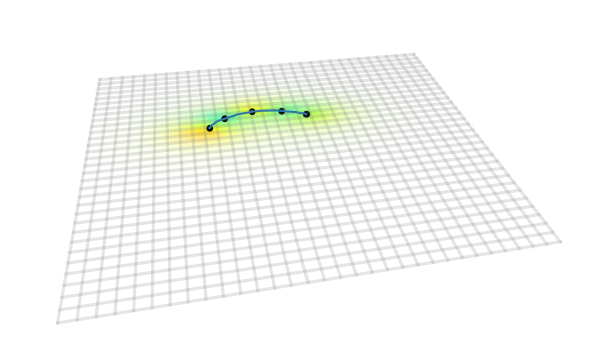} 
    \caption*{Batch 1, M = 5.}
    \end{subfigure}
    \begin{subfigure}{0.19\linewidth}
        \centering
        \includegraphics[width=\linewidth]{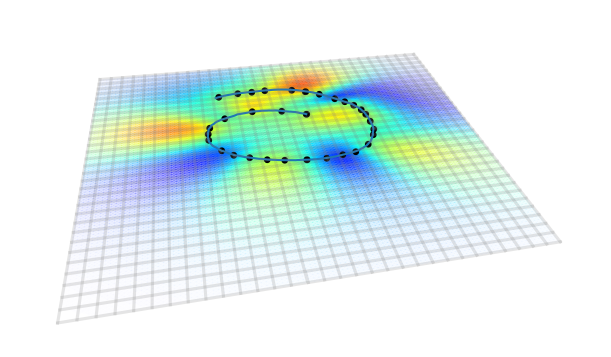}
        \caption*{Batch 5, M =  32.}
    \end{subfigure}%
    \begin{subfigure}{0.19\linewidth}
        \centering
        \includegraphics[width=\linewidth]{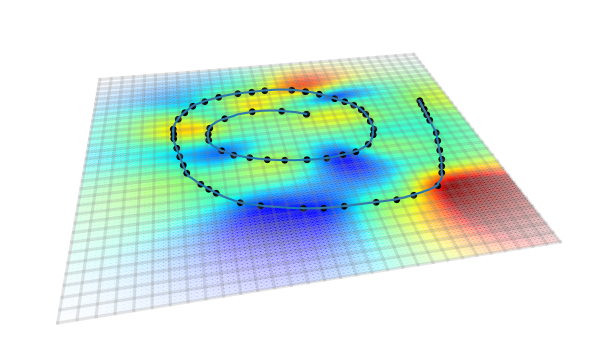} 
        \caption*{Batch 10, M = 66.}
    \end{subfigure}
    \begin{subfigure}{0.19\linewidth}
        \centering
        \includegraphics[width=\linewidth]{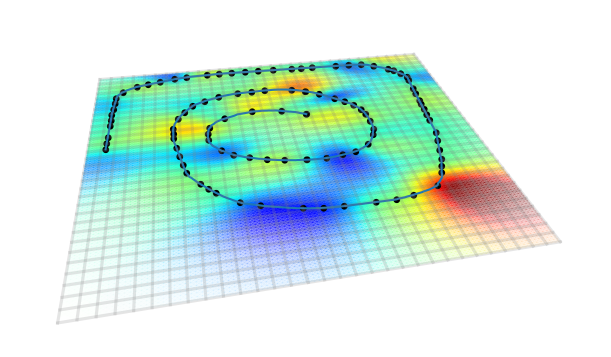} 
        \caption*{Batch 15, M = 101.}
    \end{subfigure}%
    \begin{subfigure}{0.19\linewidth}
        \centering
        \includegraphics[width=\linewidth]{figures/magneto/VIPS_model_path_3_batch_19.pdf} 
        \caption*{Batch 20, M = 134.}
    \end{subfigure}
    \caption{VIPS (Ours). A small robot with wheels is used to perform sequential estimation of magnetic field anomalies. Data is collected continuously as the robot moves along the path. The inducing points are represented as black dots and the line represents the travelled part of the path. We indicate the batch number and number of inducing points (M). Final RMSE = 7.55.}
    \label{fig:vips_magneto_path_3}
\end{center}
\end{figure}

\begin{figure}[htpb]
\begin{center}
    \begin{subfigure}{0.20\columnwidth}
        \centering
        \includegraphics[width=\linewidth]{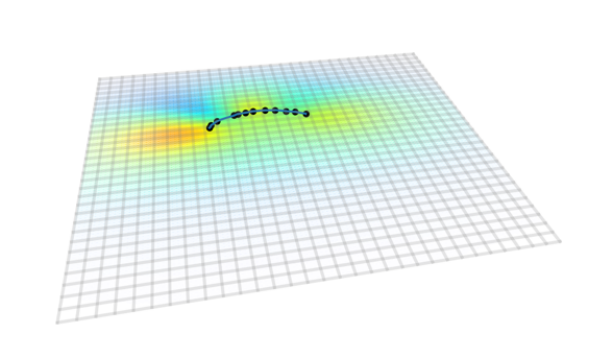} 
        \caption*{Batch 1, M = 12.}
    \end{subfigure}%
    \begin{subfigure}{0.20\columnwidth}
        \centering
        \includegraphics[width=\linewidth]{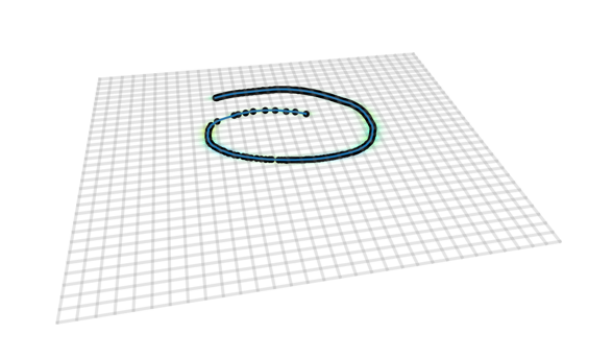} 
        \caption*{Batch 5, M = 597.}
    \end{subfigure}%
    \begin{subfigure}{0.20\columnwidth}
        \centering
        \includegraphics[width=\linewidth]{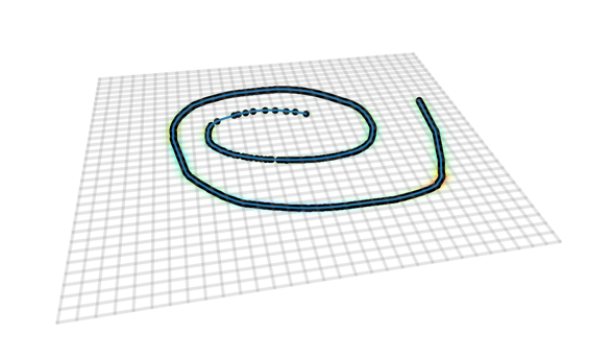} 
        \caption*{Batch 10, M = 2206.}
    \end{subfigure}%
    \begin{subfigure}{0.20\columnwidth}
        \centering
        \includegraphics[width=\linewidth]{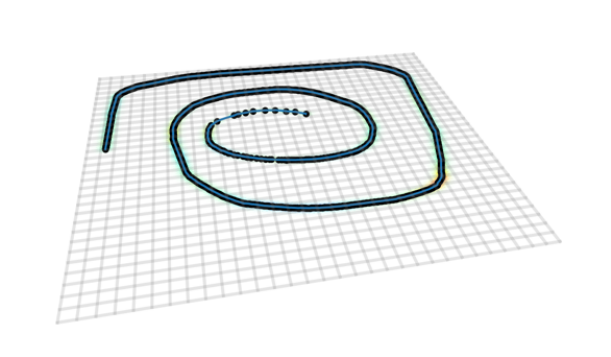} 
        \caption*{Batch 15, M = 4071.}
    \end{subfigure}%
    \begin{subfigure}{0.20\columnwidth}
        \centering
        \includegraphics[width=\linewidth]{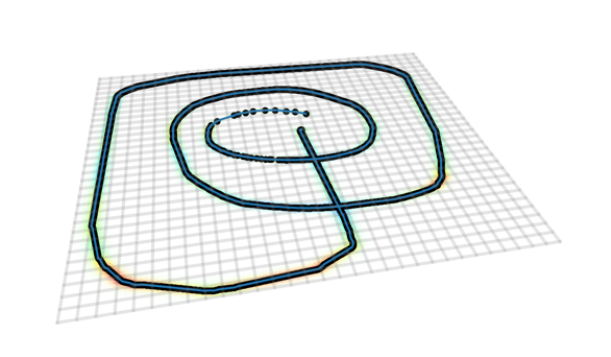} 
        \caption*{Batch 20, M = 5785.}
    \end{subfigure}
\caption{Conditional Variance. A small robot with wheels is used to perform sequential estimation of magnetic field anomalies. Data is collected continuously as the robot moves along the path. The inducing points are represented as black dots and the line represents the travelled part of the path. We indicate the batch number and number of inducing points (M). Final RMSE = 10.66.}
\label{fig:append:magneto_cv_path_3}
\end{center}
\end{figure}

\begin{figure}[htpb]
\begin{center}
    \begin{subfigure}{0.19\linewidth}
        \centering
         \includegraphics[width=\linewidth]{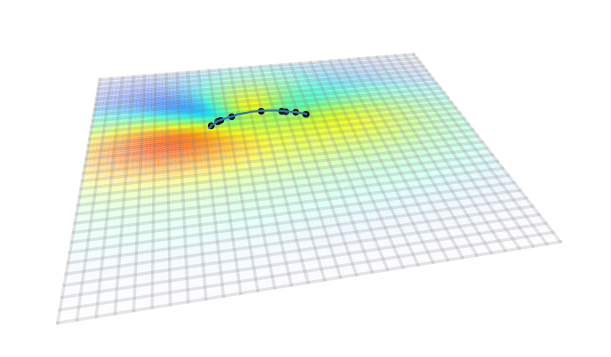}
    \caption*{Batch 1, M = 10.}
    \end{subfigure}
    \begin{subfigure}{0.19\linewidth}
        \centering
        \includegraphics[width=\linewidth]{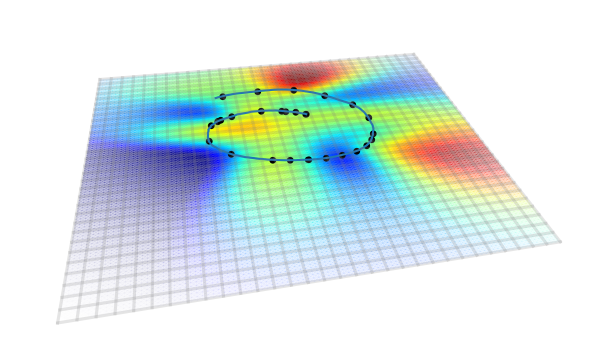}
        \caption*{Batch 5, M = 27.}
    \end{subfigure}%
    \begin{subfigure}{0.19\linewidth}
        \centering
        \includegraphics[width=\linewidth]{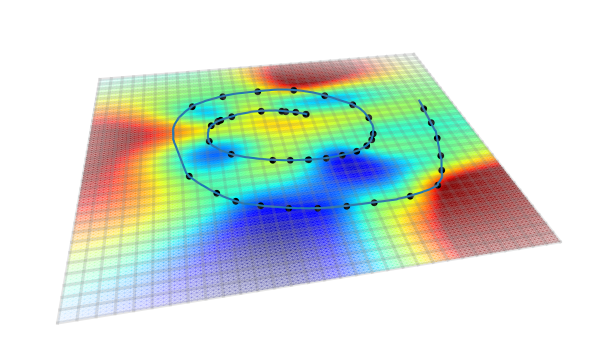} 
        \caption*{Batch 10, M = 43.}
    \end{subfigure}
    \begin{subfigure}{0.19\linewidth}
        \centering
        \includegraphics[width=\linewidth]{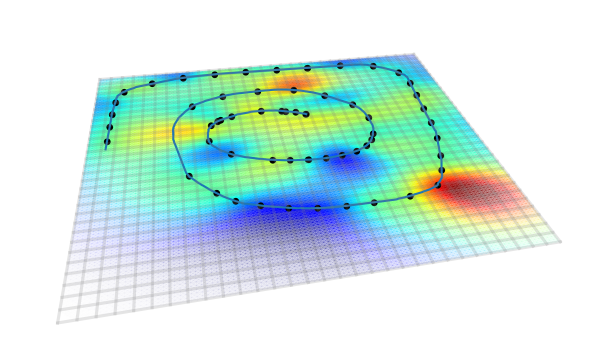} 
        \caption*{Batch 15, M = 59.}
    \end{subfigure}%
    \begin{subfigure}{0.19\linewidth}
        \centering
        \includegraphics[width=\linewidth]{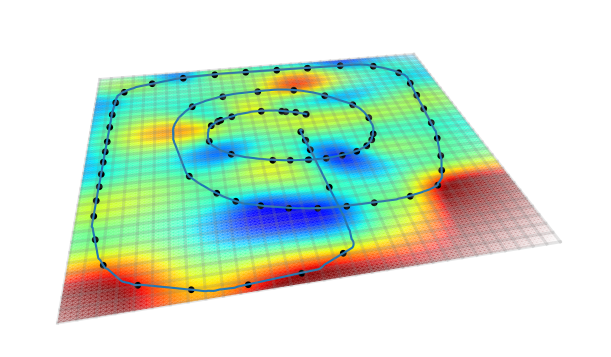} 
        \caption*{Batch 20, M = 76.}
        \end{subfigure}
    \caption{OIPS. A small robot with wheels is used to perform sequential estimation of magnetic field anomalies. Data is collected continuously as the robot moves along the path. The inducing points are represented as black dots and the line represents the travelled part of the path. We indicate the batch number and number of inducing points (M). Final RMSE = 9.42.}
    \label{fig:append:oips_magneto_path_3}
\end{center}
\end{figure}

In the second experiment, we focus on the streaming learning of trajectory 3. The trajectory is split into 20 batches. We compare the number of inducing points selected and the estimate obtained by the three methods, Conditional Variance (CV), OIPS and VIPS (ours). Detailed learning of the path for each method is shown in figures \ref{fig:vips_magneto_path_3}, \ref{fig:append:magneto_cv_path_3} and \ref{fig:append:oips_magneto_path_3}. As a test set, we use trajectories 1, 2, 4, and 5. We observed how Conditional Variance chooses an excessive number of inducing points, indicating that its hyperparameter needs tuning, which is impractical in the continual learning setting. OIPS chooses the least inducing points; as seen in the figure, the number of inducing points is more concentrated at the beginning of the path and becomes sparse towards the end. When compared to the learning estimates derived from the paths in the previous experiment, we see that both estimates differ significantly, suggesting that OIPS fails to adequately add enough capacity to capture the changes in the environment. VIPS provides the middle ground, selecting a moderate number of inducing points that effectively balance accuracy and memory size. This choice allows VIPS to maintain a robust estimate of the magnetic field obtained when compared to learning by paths without excessive computational overhead. 

In the last two sections, when compared to both alternative approaches, VIPS achieved the best trade-off between performance and model size without requiring hyperparameter tuning, making it the preferred method among the three.

\subsection{Model size and Data Distribution with Larger Data}

We reproduce the three continual learning scenarios from App.~\ref{appendix:datatypes}, increasing the dataset size to $50\,000$ data points and the input dimensionality to three, and compare how the three methods (VIPS, OIPS and Conditional Variance) scale in this setting. This setup aims to simulate practical scenarios where an exact GP would be infeasible, and a sequential approach is required instead. We use a Squared Exponential kernel initialised with lengthscale 1.0 and variance 1.0. The noise variance is set to 0.2. We evaluate all methods at the operating points corresponding to the 10\% RMSE thresholds determined from the UCI experiments (App.~\ref{appendix:UCI}). The data is divided into $50$ batches of $1000$ points each.

\paragraph{Dataset 1:} Data points are sampled uniformly from $[0, 20]^3$, with batches ordered such that new regions of the input space are discovered gradually. Model size is expected to grow linearly as more data is observed. 

\paragraph{Dataset 2:} Data points are again sampled uniformly from $[0, 20]^3$, but batches are shuffled at random. This represents a setting where the entire input space is observed from the start. As a result, the novelty of each batch decreases over time, and the model size is expected to plateau.

\paragraph{Dataset 3:} Most data is concentrated in a narrow region, with occasional outliers. Ninety per cent of the points are drawn from a truncated Cauchy distribution on $[5, 15]^3$ centred at 10, and the remainder uniformly from $[0, 40]^3$. This mimics a scenario where data is dense but includes some outliers. Therefore, the model size should remain small until the region with the outliers is observed. 

\begin{figure}[htpb]
    \centering
    \includegraphics[width=0.95\linewidth]{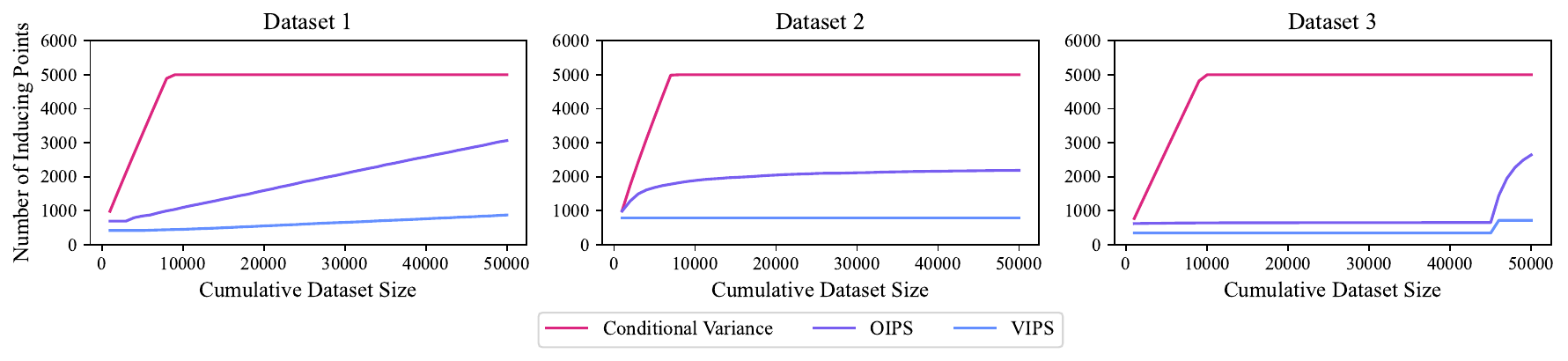}
    \caption{Number of inducing points used by each method as a function of cumulative data size, across the three synthetic datasets. VIPS consistently uses fewer inducing points and only increases model size as needed. OIPS follows a similar trend but typically uses twice as many inducing points as VIPS. CV grows unnecessarily and reaches the imposed cap of 5000 points.}
    \label{fig:data_types_large}
\end{figure}

Figure~\ref{fig:data_types_large} shows how the number of inducing points evolves as more data is observed for each method across the three datasets. VIPS adapts across all settings, increasing model size only when needed. OIPS shows similar trends but typically adds up to twice as many inducing points as VIPS. CV grows rapidly even when unnecessary (e.g. Dataset 2). Note that for CV we impose a cap of $5000$ inducing points.

The datasets were designed to be well-approximated with relatively few inducing points, and all methods attain near-exact accuracy. These results show that even at a larger scale, VIPS maintains a compact model, using fewer than a thousand inducing points in all cases. In contrast, the results for CV and OIPS suggest that these methods may require hyperparameter tuning to control model growth in larger settings.

\newpage

\section{Derivation the optimal form of variational distribution}\label{appendix:lowerbound}

The derivations follow \citet{Bui2017}. We start by rewriting the online lower bound, 
\begin{equation}
\begin{aligned}
 \widehat{\mathcal{L}} 
&= - \int \text{d}f q_{n}(f)\left[\log \frac{p(\mathbf{a} |\theta_{o})q_{n}(\mathbf{b})}{p(\mathbf{b} |\theta_{n})q_{o}(\mathbf{a})p(\mathbf{y}_{n} |f)} \right] \\
 &= \int \text{d}f  q_{n}(f) \log p(\mathbf{y}_{n}|f)
    - \int \text{d}f q_{n}(f) \log \frac{p(\mathbf{a}|\theta_{o})q_{n}(\mathbf{b})}{p(\mathbf{b}|\theta_{n})q_{o}(\mathbf{a})}.
\end{aligned}
\end{equation}

We can find the optimal $q_{n}$ by setting the derivative of $\widehat{\mathcal{L}}$ with respect to $q_{n}(\mathbf{b})$ to zero, 

\begin{equation}
\frac{\mathrm{d} \widehat{\mathcal{L}}}{\mathrm{d} q(\mathbf{b})} = \int \mathrm{d}f p\left(f_{\neq \mathbf{b}} \mid \mathbf{b}\right)\left[\log p(\mathbf{y}_{n}  \mid \mathbf{f}) - \log \frac{p\left(\mathbf{a} \mid \theta_{\text {old }}\right) q_{n}(\mathbf{b})}{p\left(\mathbf{b} \mid \theta_{\text {new }}\right) q_{o}(\mathbf{a})}\right] - 1 = 0, 
\end{equation}
this gives us,
\begin{equation*}
q_{\text {opt }}(\mathbf{b})\propto p(\mathbf{b} | \theta_{n} )\exp\left(\int \mathrm{d} \mathbf{a} p(\mathbf{a} \mid \mathbf{b}, \theta_{n}) \log \frac{q_{o}(\mathbf{a})}{p\left(\mathbf{a} \mid \theta_{\text {old }}\right)}+\int \mathrm{d} \mathbf{f} p(\mathbf{f} \mid  \mathbf{b}, \theta_{n}) \log p(\mathbf{y}_{n} \mid \mathbf{f})\right).
\end{equation*}

Let $q(\mathbf{a})=\mathcal{N}\left(\mathbf{a} ; \mathbf{m}_{\mathbf{a}}, \mathbf{S}_{\mathbf{a}}\right)$ and $p\left(\mathbf{a} \mid \theta_{\text {old }}\right)=\mathcal{N}\left(\mathbf{a} ; \mathbf{0}, \mathbf{K}_{\mathbf{aa}}^{\prime}\right)$, and denoting $\mathbf{D}_{\mathbf{a}}=\left(\mathbf{S}_{\mathbf{a}}^{-1}-\mathbf{K}_{\mathbf{a a}}^{\prime-1}\right)^{-1}, \mathbf{Q}_{\mathbf{f f }}= \mathbf{K}_{\mathbf{fb}} \mathbf{K}_{\mathbf{b b}}^{-1} \mathbf{K}_{\mathbf{b f}}$, and $\mathbf{Q}_{\mathbf{a a }}=\mathbf{K}_{\mathbf{a b}} \mathbf{K}_{\mathbf{b b}}^{-1} \mathbf{K}_{\mathbf{ba}}$, the exponents in the optimal $q_{\text {opt }}(\mathbf{b})$ can be simplified as follows:

\begin{equation}
\begin{aligned}
\mathrm{E}_1 & =\int \operatorname{d} \mathbf{a} p(\mathbf{a} \mid \mathbf{b}, \theta_{n}) \log \frac{q(\mathbf{a})}{p\left(\mathbf{a} \mid \theta_{\mathrm{o}}\right)} \\
& =\frac{1}{2} \int \mathrm{d}\mathbf{a} \mathcal{N}\left(\mathbf{a} ; \mathbf{K}_{\mathbf{a b}} \mathbf{K}_{\mathbf{b} \mathbf{b}}^{-1} \mathbf{b}, \mathbf{Q}_{\mathbf{a}}\right)\left(-\log \frac{\left|\mathbf{S}_{\mathbf{a}}\right|}{\left|\mathbf{K}_{\mathbf{a a}}^{\prime}\right|}-\left(\mathbf{a}-\mathbf{m}_{\mathbf{a}}\right)^{\top} \mathbf{S}_{\mathbf{a}}^{-1}\left(\mathbf{a}-\mathbf{m}_{\mathbf{a}}\right)+\mathbf{a}^{\top} \mathbf{K}_{\mathbf{a a}}^{\prime-1} \mathbf{a}\right) \\
& =\frac{1}{2} \int \mathrm{d}\mathbf{a} \mathcal{N}\left(\mathbf{a} ; \mathbf{K}_{\mathbf{a b}} \mathbf{K}_{\mathbf{b} \mathbf{b}}^{-1} \mathbf{b}, \mathbf{Q}_{\mathbf{a}}\right)
\left(-\log \frac{\left|\mathbf{S}_{\mathbf{a}}\right|}{\left|\mathbf{K}_{\mathbf{a a}}^{\prime}\right|}
- \mathbf{m}_{\mathbf{a}}^{\top} \mathbf{S}_{\mathbf{a}}^{-1}\mathbf{m}_{\mathbf{a}}
+  \mathbf{a}^{\top}\left(\mathbf{K}_{\mathbf{a a}}^{\prime-1} -  \mathbf{S}_{\mathbf{a a}}^{-1}\right) \mathbf{a} + 2\, \mathbf{a}^\top \mathbf{S}_{\mathbf{a}}^{-1} \mathbf{m}_{\mathbf{a}}\right)  \\
& =\frac{1}{2} \int \mathrm{d}\mathbf{a} \mathcal{N}\left(\mathbf{a} ; \mathbf{K}_{\mathbf{a b}} \mathbf{K}_{\mathbf{b} \mathbf{b}}^{-1} \mathbf{b}, \mathbf{Q}_{\mathbf{a}}\right)
\left(-\log \frac{\left|\mathbf{S}_{\mathbf{a}}\right|}{\left|\mathbf{K}_{\mathbf{a a}}^{\prime}\right|}
- \mathbf{m}_{\mathbf{a}}^{\top} \mathbf{S}_{\mathbf{a}}^{-1}\mathbf{m}_{\mathbf{a}}
-  \mathbf{a}^{\top}\mathbf{D}_{\mathbf{a}}^{-1} \mathbf{a} + 2\, \mathbf{a}^\top \mathbf{S}_{\mathbf{a}}^{-1} \mathbf{m}_{\mathbf{a}}\right) \\
&= \int \mathrm{d}\mathbf{a} \mathcal{N}\left(\mathbf{a} ; \mathbf{K}_{\mathbf{a b}} \mathbf{K}_{\mathbf{b} \mathbf{b}}^{-1} \mathbf{b}, \mathbf{Q}_{\mathbf{a}}\right)\log \,\mathcal{N}(\mathbf{D}_\mathbf{a}\mathbf{S}_\mathbf{a}^{-1}\mathbf{m}_\mathbf{a}; \mathbf{a},  \mathbf{D}_\mathbf{a}) + \Delta_\mathbf{a}\\
& =\log \mathcal{N}\left(\mathbf{D}_{\mathbf{a}} \mathbf{S}_{\mathbf{a}}^{-1} \mathbf{m}_{\mathbf{a}} ; \mathbf{K}_{\mathbf{a b}} \mathbf{K}_{\mathbf{b}}^{-1} \mathbf{b}, \mathbf{D}_{\mathbf{a}}\right) -\operatorname{tr}\left[\mathbf{D}_{\mathbf{a}}^{-1} \left( \mathbf{K}_{\mathbf{a a }} - \mathbf{Q}_{\mathbf{a a }}\right)\right] +\Delta_\mathbf{a} \\
\mathrm{E}_2 & =\int \mathrm{d} \mathbf{f} p(\mathbf{f} \mid \mathbf{b}, \theta_{n}) \log p(\mathbf{y}_{n} \mid \mathbf{f}) \\
& =\int \mathrm{d} \mathbf{f} \mathcal{N}\left(\mathbf{f} ; \mathbf{K}_{\mathbf{f b}} \mathbf{K}_{\mathbf{b}}^{-1} \mathbf{b}, \mathbf{Q}_{\mathbf{f}}\right) \log \mathcal{N}\left(\mathbf{y} ; \mathbf{f}, \sigma^2 \mathrm{I}\right) \\
& =\log \mathcal{N}\left(\mathbf{y} ; \mathbf{K}_{\mathbf{f b}} \mathbf{K}_{\mathbf{b b}}^{-1} \mathbf{b}, \sigma^2 \mathrm{I}\right) -\frac{1}{2 \sigma^2} \operatorname{tr}\left(\mathbf{K}_{\mathbf{f f }} - \mathbf{Q}_{\mathbf{f f}}\right), \\
2 \Delta_\mathbf{a} & =-\log \frac{\left|\mathbf{S}_{\mathbf{a}}\right|}{\left|\mathbf{K}_{\mathbf{a a}}^{\prime}\right|\left|\mathbf{D}_{\mathbf{a}}\right|}+\mathbf{m}_{\mathbf{a}}^{\top} \mathbf{S}_{\mathbf{a}}^{-1} \mathbf{D}_{\mathbf{a}} \mathbf{S}_{\mathbf{a}}^{-1} \mathbf{m}_{\mathbf{a}} -\mathbf{m}_{\mathbf{a}}^{\top} \mathbf{S}_{\mathbf{a}}^{-1} \mathbf{m}_{\mathbf{a}}+M_{\mathbf{a}} \log (2 \pi), \\
\end{aligned}
\end{equation}
where $\Delta_\mathbf{a}$ is a constant term with respesct to $\mathbf{b}$ and $M_{\mathbf{a}}$ is the dimension of $\mathbf{a}$. The optimal form of $q_{n}$ is then given by:
\begin{equation}
q_{\mathrm{opt}}(\mathbf{b} ) \propto p(\mathbf{b}|\theta_{n})\mathcal{N}\left(\hat{\mathbf{y}}, \mathbf{K}_{\hat{\mathbf{f}} \mathbf{b}} \mathbf{K}_{\mathbf{b} \mathbf{b}}^{-1} \mathbf{b}, \Sigma_{\hat{\mathbf{y}}}\right)
\end{equation}
where
\begin{equation}
\hat{\mathbf{y}}=\left[\begin{array}{c}
\mathbf{y} \\
\mathbf{D}_{\mathbf{a}} \mathbf{S}_{\mathbf{a}}^{-1} \mathbf{m}_{\mathbf{a}}
\end{array}\right], \mathbf{K}_{\hat{\mathbf{f b}}}=\left[\begin{array}{c}
\mathbf{K}_{\mathrm{fb}} \\
\mathbf{K}_{\mathrm{ab}}
\end{array}\right], \Sigma_{\hat{\mathbf{y}}}=\left[\begin{array}{cc}
\sigma_y^2 \mathrm{I} & \mathbf{0} \\
\mathbf{0} & \mathbf{D}_{\mathbf{a}}
\end{array}\right] .
\end{equation}
After normalisation, we have 
\begin{equation}
    q_{\mathrm{opt}}(\mathbf{b} ) =  \frac{p(\mathbf{b}|\theta_{n})\mathcal{N}\left(\hat{\mathbf{y}}, \mathbf{K}_{\hat{\mathbf{f}} \mathbf{b}} \mathbf{K}_{\mathbf{b} \mathbf{b}}^{-1} \mathbf{b}, \Sigma_{\hat{\mathbf{y}}}\right)}{\int  \mathrm{d} \mathbf{b}  p(\mathbf{b}|\theta_{n})\mathcal{N}\left(\hat{\mathbf{y}}, \mathbf{K}_{\hat{\mathbf{f}} \mathbf{b}} \mathbf{K}_{\mathbf{b} \mathbf{b}}^{-1} \mathbf{b}, \Sigma_{\hat{\mathbf{y}}}\right)}
\end{equation}

Substituting the above results into the lower bound, we have
\begin{equation}
    \widehat{\mathcal{L}} =\log \mathcal{N}\left(\hat{\mathbf{y}} ; \mathbf{0}, \mathbf{K}_{\hat{\mathbf{f}} \mathbf{b}} \mathbf{K}_{\mathbf{b b}}^{-1} \mathbf{K}_{\mathbf{b} \hat{\mathbf{f}}}+\Sigma_{\hat{\mathbf{y}}}\right)  + \Delta_\mathbf{a} - \frac{1}{2} \operatorname{tr}\left[\mathbf{D}_\mathbf{a}^{-1}(\mathbf{K}_\mathbf{aa} - \mathbf{Q}_\mathbf{aa} )  \right]- \frac{1}{2\sigma^2} \operatorname{tr}(\mathbf{K}_\mathbf{ff} - \mathbf{Q}_\mathbf{ff} ).
\end{equation}

\subsection{Derivation of $\mathcal{L}^*$}

Recall that 
\begin{equation}
    \mathcal{L}^* =  \log \mathcal{N} \left(
            \hat{\mathbf{y}} ;\,\mathbf{0}\,,  \mathbf{K}_{\hat{\mathbf{f}}\hat{\mathbf{f}}} + \Sigma_{\hat{y}}\right)  +  \Delta_\mathbf{a}, 
\end{equation}
with $\mathbf{K}_{\hat{\mathbf{f}}\hat{\mathbf{f}}} = \left[\begin{array}{cc}
                \mathbf{K}_{\mathbf{ff}} & \mathbf{K}_{\mathbf{fa}} \\
                \mathbf{K}_{\mathbf{af}} & \mathbf{K}_{\mathbf{aa}}
                \end{array}\right].$

The first term can be lower bounded using Jensen's inequality as, 
\begin{equation}
\begin{aligned}
    &\log \mathcal{N} \left(
            \hat{\mathbf{y}} ;\,\mathbf{0}\,,  \mathbf{K}_{\hat{\mathbf{f}}\hat{\mathbf{f}}} + \Sigma_{\hat{y}}\right) \\
    &\quad\geq \log \mathcal{N} \left(
            \hat{\mathbf{y}} ;\,\mathbf{0}\,,   \mathbf{K}_{\mathbf{\hat{f}b}}\mathbf{K}_{\mathbf{bb}}^{-1}\mathbf{K}_{\mathbf{b\hat{f}}} +  \Sigma_{\mathbf{\hat{y}}} \right) - \frac{1}{2} \operatorname{tr}\left( \Sigma_{\mathbf{\hat{y}}}^{-1} \left( \mathbf{K}_{\hat{\mathbf{f}}\hat{\mathbf{f}}} - \mathbf{K}_{\mathbf{\hat{f}b}}\mathbf{K}_{\mathbf{bb}}^{-1}\mathbf{K}_{\mathbf{b\hat{f}}} \right) \right)
\end{aligned}
\end{equation}
where $\mathbf{K}_{\mathbf{\hat{f}b}}\mathbf{K}_{\mathbf{bb}}^{-1}\mathbf{K}_{\mathbf{b\hat{f}}} $ is the Nyström approximation of $\mathbf{K}_{\mathbf{\hat{f}\hat{f}}}$. The trace will be small when $\mathbf{b} = \{f(\mathbf{x}_{n}, \mathbf{a})\}$ and can be simplified as follows:
\begin{equation}
\begin{aligned}
 &\operatorname{tr}\left( \Sigma_{\mathbf{\hat{y}}}^{-1} \left(  \mathbf{K}_{\mathbf{\hat{f}\hat{f}}} - \mathbf{K}_{\mathbf{\hat{f}b}}\mathbf{K}_{\mathbf{bb}}^{-1}\mathbf{K}_{\mathbf{b\hat{f}}} \right) \right) \\[5pt]
&\quad = \operatorname{tr}\left( \left[\begin{array}{cc}
\sigma_y^{-2} \mathbf{I} & \mathbf{0} \\
\mathbf{0} & \mathbf{D}_{\mathbf{a}}^{-1}
\end{array}\right] \left( \left[\begin{array}{cc}
                \mathbf{K}_{\mathbf{ff}} & \mathbf{K}_{\mathbf{fa}} \\
                \mathbf{K}_{\mathbf{af}} & \mathbf{K}_{\mathbf{aa}}
                \end{array}\right] - \left[\begin{array}{cc}
                \mathbf{K}_{\mathbf{fb}}\mathbf{K}_{\mathbf{bb}}^{-1}\mathbf{K}_{\mathbf{bf}} & \mathbf{K}_{\mathbf{fb}}\mathbf{K}_{\mathbf{bb}}^{-1}\mathbf{K}_{\mathbf{ba}} \\[2pt]
                \mathbf{K}_{\mathbf{ab}}\mathbf{K}_{\mathbf{bb}}^{-1}\mathbf{K}_{\mathbf{bf}} & \mathbf{K}_{\mathbf{ab}}\mathbf{K}_{\mathbf{bb}}^{-1}\mathbf{K}_{\mathbf{ba}}
                \end{array}\right] \right) \right) \\[5pt]
&\quad = 
    \operatorname{tr}\left(\mathbf{D}_\mathbf{a}^{-1}(\mathbf{K}_\mathbf{aa} - \mathbf{K}_{\mathbf{ab}}\mathbf{K}_{\mathbf{bb}}^{-1}\mathbf{K}_{\mathbf{ba}} )  \right) +  \frac{1}{\sigma^2} \operatorname{tr}(\mathbf{K}_\mathbf{ff} - \mathbf{K}_{\mathbf{fb}}\mathbf{K}_{\mathbf{bb}}^{-1}\mathbf{K}_{\mathbf{bf}} ) 
\end{aligned}
\end{equation}
which recovers the expression for  $\widehat{\mathcal{L}} - \Delta_\mathbf{a}$ .

\section{Online Upper Bound Implementation}\label{appendix:upperbound}

In this section, we provide efficient forms for practical implementation of the online upper bound  $\mathcal{\widehat{U}}$. As the second term is constant we focus on the first term, 
\begin{equation}
 \mathcal{\widehat{U}}_2 = -\frac{(N+ M_\mathbf{a})}{2} \log(2\pi)
- \frac{1}{2} \log  |\mathbf{K}_{\mathbf{\hat{f}b}}\mathbf{K}_{\mathbf{bb}}^{-1}\mathbf{K}_{\mathbf{b\hat{f}}} + \Sigma_{\mathbf{\hat{y}}} | - \frac{1}{2} \hat{\mathbf{y}}^T\left( \mathbf{K}_{\mathbf{\hat{f}b}}\mathbf{K}_{\mathbf{bb}}^{-1}\mathbf{K}_{\mathbf{b\hat{f}}} +t\mathbf{I}+ \Sigma_{\mathbf{\hat{y}}}\right)^{-1}\hat{\mathbf{y}}.
\end{equation}
This term is an upper bound for the first term of 
$\mathcal{L}^* =  \log \mathcal{N} \left(
            \hat{\mathbf{y}} ;\,\mathbf{0}\,, \mathbf{K}_{\hat{\mathbf{f}}\hat{\mathbf{f}}} + \Sigma_{\hat{y}}\right)  +  \Delta_\mathbf{a}$.

\subsection{Determinant term}
Letting $\mathbf{K}_{\mathbf{bb}} =\mathbf{L}_\mathbf{b}\mathbf{L}_\mathbf{b}^T$ and using the matrix determinant lemma, we can rewrite the determinant term as

\begin{equation}
\begin{aligned}
\log  |\mathbf{K}_{\mathbf{\hat{f}b}}\mathbf{K}_{\mathbf{bb}}^{-1}\mathbf{K}_{\mathbf{b\hat{f}}} + \Sigma_{\mathbf{\hat{y}}} |  =& \log  |\Sigma_{\mathbf{\hat{y}}} | + \log | \mathbf{I} + \mathbf{L}_\mathbf{b}^{-1}\mathbf{K}_{\mathbf{b\hat{f}}} \Sigma_{\mathbf{\hat{y}}}^{-1}\mathbf{K}_{\mathbf{\hat{f}b}} \mathbf{L}_\mathbf{b}^{-T} | \\
=& N\log \sigma^2_y + \log  |\mathbf{D}_\mathbf{a} | + \log | \mathbf{I} + \mathbf{L}_\mathbf{b}^{-1}\mathbf{K}_{\mathbf{b\hat{f}}} \Sigma_{\mathbf{\hat{y}}}^{-1}\mathbf{K}_{\mathbf{\hat{f}b}} \mathbf{L}_\mathbf{b}^{-T} | 
\end{aligned}
\end{equation}

Let $\mathbf{D} = \mathbf{I} + \mathbf{L}_\mathbf{b}^{-1}\mathbf{K}_{\mathbf{b\hat{f}}} \Sigma_{\mathbf{\hat{y}}}^{-1}\mathbf{K}_{\mathbf{\hat{f}b}} \mathbf{L}_\mathbf{b}^{-T}$. Note that, 

\begin{equation}
\begin{aligned}
\mathbf{K}_{\mathbf{b\hat{f}}} \Sigma_{\mathbf{\hat{y}}}^{-1}\mathbf{K}_{\mathbf{\hat{f}b}} &= 
\begin{bmatrix} 
\mathbf{K}_{\mathbf{bf}}& \mathbf{K}_\mathbf{ba}\\
\end{bmatrix}
\begin{bmatrix} 
\frac{1}{\sigma^2_y}\mathbf{I} & 0 \\
0 & \mathbf{D}_\mathbf{a}^{-1}\\
\end{bmatrix} 
\begin{bmatrix} 
\mathbf{K}_{\mathbf{fb}}\\
\mathbf{K}_{\mathbf{ab}}\\
\end{bmatrix} \\
&= \frac{1}{\sigma^2_y}\mathbf{K}_{\mathbf{bf}}\mathbf{K}_{\mathbf{fb}}  + \mathbf{K}_{\mathbf{ba}}\mathbf{D}_\mathbf{a}^{-1}\mathbf{K}_{\mathbf{ab}} \\
&= \frac{1}{\sigma^2_y}\mathbf{K}_{\mathbf{bf}}\mathbf{K}_{\mathbf{fb}}  + \mathbf{K}_{\mathbf{ba}}\mathbf{S}_\mathbf{a}^{-1}\mathbf{K}_{\mathbf{ab}} - \mathbf{K}_{\mathbf{ba}} \mathbf{K}_\mathbf{aa}^{'-1}\mathbf{K}_{\mathbf{ab}}.
\end{aligned}
\end{equation}
Therefore, 

\begin{equation}
   \mathbf{D} =  \mathbf{I} +  \frac{1}{\sigma^2_y}\mathbf{L}_\mathbf{b}^{-1}\mathbf{K}_{\mathbf{bf}}\mathbf{K}_{\mathbf{fb}}\mathbf{L}_\mathbf{b}^{-T}  + \mathbf{L}_\mathbf{b}^{-1}\mathbf{K}_{\mathbf{ba}}\mathbf{S}_\mathbf{a}^{-1}\mathbf{K}_{\mathbf{ab}}\mathbf{L}_\mathbf{b}^{-T} - \mathbf{L}_\mathbf{b}^{-1}\mathbf{K}_{\mathbf{ba}} \mathbf{K}_\mathbf{aa}^{'-1}\mathbf{K}_{\mathbf{ab}}\mathbf{L}_\mathbf{b}^{-T}.
\end{equation}

\subsection{Quadratic term}
Given the quadratic term, 
\begin{equation*}
    - \frac{1}{2} \hat{\mathbf{y}}^T\left( \mathbf{K}_{\mathbf{\hat{f}b}}\mathbf{K}_{\mathbf{bb}}^{-1}\mathbf{K}_{\mathbf{b\hat{f}}} +t\mathbf{I}+ \Sigma_{\mathbf{\hat{y}}} \right)^{-1}\hat{\mathbf{y}}.
\end{equation*}
Letting $\widehat{\Sigma}_{\mathbf{\hat{y}}} = t\mathbf{I} + \Sigma_{\mathbf{\hat{y}}}$ and by Woodbury's formula, we obtain:

\begin{equation*}
    \left( \mathbf{K}_{\mathbf{\hat{f}b}}\mathbf{K}_{\mathbf{bb}}^{-1}\mathbf{K}_{\mathbf{b\hat{f}}} + \widehat{\Sigma}_{\mathbf{\hat{y}}} \right)^{-1} = \widehat{\Sigma}_{\mathbf{\hat{y}}}^{-1} - 
\widehat{\Sigma}_{\mathbf{\hat{y}}}^{-1}\mathbf{K}_{\mathbf{\hat{f}b}}\mathbf{L}_\mathbf{b}^{-T}\left(I +  \mathbf{L}_\mathbf{b}^{-1}\mathbf{K}_{\mathbf{b\hat{f}}}\widehat{\Sigma}_{\mathbf{\hat{y}}}^{-1}\mathbf{K}_{\mathbf{\hat{f}b}}\mathbf{L}_\mathbf{b}^{-T}\right)^{-1} \mathbf{L}_\mathbf{b}^{-1}\mathbf{K}_{\mathbf{b\hat{f}}}\widehat{\Sigma}_{\mathbf{\hat{y}}}^{-1}.
\end{equation*}
We have, 
\begin{equation}
  \hat{\mathbf{y}}^T\widehat{\Sigma}_{\mathbf{\hat{y}}}^{-1}\hat{\mathbf{y}} = \frac{1}{\sigma^2_y + t}\mathbf{y}^T\mathbf{y} + \left(\mathbf{D}_\mathbf{a}\mathbf{S}_\mathbf{a}^{-1}\mathbf{m}_\mathbf{a}\right)^T\left(\mathbf{D}_\mathbf{a} + tI\right)^{-1}\left(\mathbf{D}_\mathbf{a}\mathbf{S}_\mathbf{a}^{-1}\mathbf{m}_\mathbf{a}\right).
\end{equation}
Letting $\widehat{\mathbf{D}} =\mathbf{I}+  \mathbf{L}_\mathbf{b}^{-1}\mathbf{K}_{\mathbf{b\hat{f}}}\widehat{\Sigma}_{\mathbf{\hat{y}}}^{-1}\mathbf{K}_{\mathbf{\hat{f}b}}\mathbf{L}_\mathbf{b}^{-T}$ where,  
\begin{align*}
\mathbf{K}_{\mathbf{b\hat{f}}} \Sigma_\mathbf{c}^{-1}\mathbf{K}_{\mathbf{\hat{f}b}} &= 
\begin{bmatrix}^\top 
\mathbf{K}_{\mathbf{fb}}\\
\mathbf{K}_{\mathbf{ab}}\\
\end{bmatrix} 
\begin{bmatrix} 
\frac{1}{\sigma^2_y + t}\mathbf{I} & 0 \\
0 & (\mathbf{D}_\mathbf{a} + t\mathbf{I})^{-1}\\
\end{bmatrix} 
\begin{bmatrix} 
\mathbf{K}_{\mathbf{fb}}\\
\mathbf{K}_{\mathbf{ab}}\\
\end{bmatrix} 
= \frac{1}{\sigma^2_y + t}\mathbf{K}_{\mathbf{bf}}\mathbf{K}_{\mathbf{fb}}  + \mathbf{K}_{\mathbf{ba}}(\mathbf{D}_\mathbf{a} + t\mathbf{I})^{-1}\mathbf{K}_{\mathbf{ab}} 
\end{align*}
and letting $ \hat{\mathbf{c}}= \mathbf{K}_{\mathbf{b\hat{f}}}\widehat{\Sigma}_{\mathbf{\hat{y}}}^{-1}\hat{\mathbf{y}} 
  = \frac{1}{\sigma^2_y + t}\mathbf{K}_{\mathbf{bf}}\mathbf{y} + \mathbf{K}_{\mathbf{ba}}\left(\mathbf{D}_\mathbf{a} + tI\right)^{-1}\left(\mathbf{D}_\mathbf{a}\mathbf{S}_\mathbf{a}^{-1}\mathbf{m}_\mathbf{a}\right)$, 
we obtain, 
\begin{equation}
    \hat{\mathbf{y}} ^T\widehat{\Sigma}_{\mathbf{\hat{y}}}^{-1}\mathbf{K}_{\mathbf{\hat{f}b}}\mathbf{L}_\mathbf{b}^{-T}\left(I +  \mathbf{L}_\mathbf{b}^{-1}\mathbf{K}_{\mathbf{b\hat{f}}}\widehat{\Sigma}_{\mathbf{\hat{y}}}^{-1}\mathbf{K}_{\mathbf{\hat{f}b}}\mathbf{L}_\mathbf{b}^{-T}\right)^{-1} \mathbf{L}_\mathbf{b}^{-1}\mathbf{K}_{\mathbf{b\hat{f}}}\widehat{\Sigma}_{\mathbf{\hat{y}}}^{-1}\hat{\mathbf{y}}  = \hat{\mathbf{c}}^T\mathbf{L}_\mathbf{b}^{-T}\widehat{\mathbf{D}}^{-1}\mathbf{L}_\mathbf{b}^{-1}\hat{\mathbf{c}}.
\end{equation}
Putting this back into the upper bound: 

\begin{equation}
\mathcal{\widehat{U}}_2 =  -\frac{(N+ M_\mathbf{a})}{2} \log(2\pi)
- \frac{1}{2}N\log \sigma^2_y - \frac{1}{2}\log  |\mathbf{D}_\mathbf{a} | - \frac{1}{2} \log |\mathbf{D} | - \frac{1}{2}\hat{\mathbf{y}}^T\widehat{\Sigma}_{\mathbf{\hat{y}}}^{-1}\hat{\mathbf{y}} + \frac{1}{2}\hat{\mathbf{c}}^T\mathbf{L}_\mathbf{b}^{-T}\widehat{\mathbf{D}}^{-1}\mathbf{L}_\mathbf{b}^{-1}\hat{\mathbf{c}}.
\end{equation}
The upper bound for $\mathcal{L}^*$ is therefore
\begin{equation}
\begin{aligned}
\mathcal{\widehat{U}} 
&= -\frac{(N+ M_\mathbf{a})}{2} \log(2\pi)
- \frac{1}{2} \log  |\mathbf{K}_{\mathbf{\hat{f}b}}\mathbf{K}_{\mathbf{bb}}^{-1}\mathbf{K}_{\mathbf{b\hat{f}}} + \Sigma_{\mathbf{\hat{y}}} | - \frac{1}{2} \hat{\mathbf{y}}^T\left( \mathbf{K}_{\mathbf{\hat{f}b}}\mathbf{K}_{\mathbf{bb}}^{-1}\mathbf{K}_{\mathbf{b\hat{f}}} +t\mathbf{I}+ \Sigma_{\mathbf{\hat{y}}} \right)^{-1}\hat{\mathbf{y}} + \Delta_\mathbf{a} \\
&=   -\frac{N}{2}\log(2\pi \sigma_y^2) - \frac{1}{2} \log |\mathbf{D} | - \frac{1}{2}\frac{1}{\sigma^2_y + t}\mathbf{y}^T\mathbf{y} - \frac{1}{2} \left(\mathbf{D}_\mathbf{a}\mathbf{S}_\mathbf{a}^{-1}\mathbf{m}_\mathbf{a}\right)^T\left(\mathbf{D}_\mathbf{a} + tI\right)^{-1}\left(\mathbf{D}_\mathbf{a}\mathbf{S}_\mathbf{a}^{-1}\mathbf{m}_\mathbf{a}\right) \\
&\quad+ \frac{1}{2}\hat{\mathbf{c}}^T\mathbf{L}_\mathbf{b}^{-T}\widehat{\mathbf{D}}^{-1}\mathbf{L}_\mathbf{b}^{-1}\hat{\mathbf{c}}  -\frac{1}{2} \log \frac{ |\mathbf{S}_\mathbf{a} |}{ |\mathbf{K}_\mathbf{aa}^\prime |} -\frac{1}{2}\mathbf{m}_\mathbf{a}^T\mathbf{S}_\mathbf{a}^{-1}\mathbf{m}_\mathbf{a} + \frac{1}{2}\mathbf{m}_\mathbf{a}^T\mathbf{S}_\mathbf{a}^{-1}\mathbf{D}_\mathbf{a}\mathbf{S}_\mathbf{a}^{-1}\mathbf{m}_\mathbf{a}.
\end{aligned}
\end{equation}

\end{document}